\documentclass{article}
\usepackage{times}
\usepackage[margin=1in]{geometry}
\usepackage{graphicx} 
\usepackage{subfigure}

\usepackage{algorithm}
\usepackage{algorithmic}
\usepackage{amsmath}
\usepackage{natbib}
\usepackage{hyperref}
\usepackage{amssymb}
\usepackage{amsthm}

\newcommand*{\QEDB}{\hfill\ensuremath{\square}}

\newtheorem{assumption}{\textbf{Assumption}}
\newtheorem{theorem}{\textbf{Theorem}}
\newtheorem*{theorem*}{\textbf{Theorem}}
\newtheorem{lemma}{\textbf{Lemma}}
\newtheorem{corollary}{\textbf{Corollary}}
\newtheorem{proposition}{\textbf{Proposition}}
\newtheorem*{proposition*}{\textbf{Proposition}}

\title{Fast Stochastic Variance Reduced ADMM for Stochastic Composition Optimization}
\author{Yue Yu and Longbo Huang\\
Institute for Interdisciplinary Information Sciences, Tsinghua University  \\
yu-y14@mails.tsinghua.edu.cn, longbohuang@tsinghua.edu.cn}

\begin{document}

\maketitle

\begin{abstract}
We consider the  stochastic composition optimization problem  proposed in \cite{wang2017stochastic}, which  has applications ranging from estimation to statistical and machine learning. We propose the first ADMM-based algorithm named com-SVR-ADMM, and show that com-SVR-ADMM converges linearly for  strongly convex and Lipschitz smooth objectives, and has a convergence rate of $O( \log S/S)$, which improves upon the $O(S^{-4/9})$ rate in \cite{wang2016accelerating} when the objective is convex and Lipschitz smooth. 
Moreover, com-SVR-ADMM possesses a rate of $O(1/\sqrt{S})$ when the objective is convex but without Lipschitz smoothness.  We also conduct experiments  and show that it outperforms existing algorithms.
\end{abstract}

\section{Introduction}
Recently, \cite{wang2017stochastic} proposed the  stochastic composition optimization of the following form:
\begin{equation}
  \min_{x}(\mathbf{E}_if_i \circ \mathbf{E}_j g_j)(x). \label{eq:problem1}
\end{equation}
Here $x \in \mathbb{R}^{q}$, $\mathbf{E}_if_i = \mathbf{E}_i\{f_i(x)  \}$, $(f\circ g) (x) \triangleq f(g(x))$ denotes the composite function, and $i, j$ are random variables. 
%
Problem (\ref{eq:problem1}) has been shown in \cite{wang2017stochastic} to include several important applications in estimation and machine learning.

In this paper, we focus on extending the formulation to include linear constraints, and consider the following variant of Problem (\ref{eq:problem1}):
\begin{eqnarray}
&\min_{x,\omega}& F(x) + R(\omega) \label{eq:problem-2-obj}\\
&\text{s.t.}&  Ax+B\omega = 0. \label{eq:problem-2-con}
\end{eqnarray}
Here $F(x)\triangleq\frac{1}{n} \sum\limits_{i=1}^{n} f_{i}(\frac{1}{m}\sum\limits_{j=1}^{m}g_j(x))$, $x \in \mathbb{R}^{q}$, $\omega \in \mathbb{R}^{l}$, $A \in \mathbb{R}^{p \times q}$, $B\in \mathbb{R}^{p\times l}$, $g_j: \mathbb{R}^{q} \mapsto \mathbb{R}^{r}$ and $f_{i}: \mathbb{R}^{r} \mapsto \mathbb{R}$ are continuous functions, and  $R(\omega): \mathbb{R}^{l} \mapsto \mathbb{R}$ is a closed convex function. %
The reason to consider the specific form of Problem (\ref{eq:problem-2-obj}) is as follows. (i) In practice, random variables such as $i$ and $j$ are obtained from problem-dependent data sets. Thus, they often only take values in a finite set with certain frequencies (captured by the first term in the objective (\ref{eq:problem-2-obj})).  (ii) Such problems often require the solutions to satisfy certain regularizing conditions (imposed by the term $R(\omega)$ and constraint (\ref{eq:problem-2-con})).
Note here that  the uniform distribution of  $i$ and $j$ (the $\frac{1}{n}$ and $\frac{1}{m}$) in (\ref{eq:problem-2-obj}) is not critical. In Section \ref{section:simulation}, we show that our algorithm is also applicable under other distributions.



\subsection{Motivating Examples}
We first present a few motivating examples of formulation (\ref{eq:problem-2-obj}). The first example is a risk-averse learning problem discussed in \cite{wang2016accelerating}, which can be formulated into the following mean-variance minimization problems,  i.e.,
\begin{eqnarray}
 \label{eq:mean-var}
&\min_{x}& \mathbf{E}_{\epsilon} h_{\epsilon}(x) + \lambda \textbf{Var}_{\epsilon}h_{\epsilon}(x) \\
&\text{s.t.}& Ax =0.
\end{eqnarray}
Here $h_{\epsilon}(x)$ is the loss function w.r.t variable $x$ and is parameterized by random variable $\epsilon$, and  $\textbf{Var}_\epsilon(x)\triangleq  \mathbf{E}_{\epsilon} \{[h_\epsilon(x) -  \mathbf{E}_{\epsilon} h_{\epsilon}(x)]^2\}$ denotes its variance.
We see that this example is of the form (\ref{eq:problem-2-obj}), where $ \mathbf{E}_{\epsilon} h_{\epsilon}(x)$ plays the role of the regularizer and the variance term is the composition functions. There are many other problems that can be formulated into the mean-variance optimization as in (\ref{eq:mean-var}), e.g., portfolio management \cite{alexander2004comparison}.

The second motivating example is dynamic programming \cite{sutton1998reinforcement,dai2016learning}.
In this case, one can often approximate the value of each state by an inner product of a state feature $\phi_s$ and a target variable $w$. Then, the policy learning problem can be formulated into minimizing the Bellman residual as follows:
\begin{equation}
  \label{eq:rl-bell}
  \min_{w}\sum\limits_{s=1}^{S}(\langle\,\phi_s, w\rangle - \sum_{s'} P_{s,s'}^{\pi}(r_{s,s'} + \gamma \langle\,\phi_{s'},w\rangle))^2 + R(w),
\end{equation}
where $P_{s, s'}^{\pi}$ denotes the transition probabilities under a policy $\pi$, and $\gamma$ denotes the discounting factor.
Note that this problem also has the form of  Problem (\ref{eq:problem-2-obj}).

The third example is  multi-stage stochastic programming  \cite{shapiro2014lectures}. For example, a two-stage optimization scenario often requires solving the following problem:
\begin{equation*}
\min_x \mathbf{E}_{v}(\min_{y} \mathbf{E}_{u|v}(U(x, v, y, u))).
\end{equation*}
Here $x, y$ are decision variables, $v, u$ are the corresponding random variables, and the function $U$ is the utility function. In this case, the  expectation $\mathbf{E}_{u|v}(\cdot)$ is the inner function and $\min_y(\cdot)$ is the outer function in Problem (\ref{eq:problem-2-obj}).

From these examples, we see that  formulation (\ref{eq:problem-2-obj}) is general and includes  important applications. Thus, it is important to develop fast and robust algorithms for solving (\ref{eq:problem-2-obj}).


\subsection{Related Works}
The stochastic composition optimization problem was first proposed in  \cite{wang2017stochastic}, where two solution algorithms, Basic SCGD and accelerated SCGD, were proposed. The algorithms were shown to achieve a sublinear convergence rate for convex and strongly convex cases, and  were also shown to converge to a stationary point in the nonconvex case.
Later, \cite{wang2016accelerating} proposed a proximal gradient algorithm called ASC-PG to improve the convergence rate when both inner and outer functions are smooth. However, the convergence rate is sublinear and their results do not include the regularizer when the objective functions are not strongly convex.
 In  \cite{lian2016finite}, the authors solved the finite sample case of stochastic composition optimization and obtained two linear-convergent algorithms based on the stochastic variance reduction gradient technique
(SVRG) proposed in \cite{johnson2013accelerating}. However,  the algorithms do not handle the regularizer either.
 %


The ADMM algorithm, on the other hand, was first proposed in \cite{glowinski1975approximation,gabay1976dual} and later reviewed in
\cite{boyd2011distributed}. Since then, several ADMM-based stochastic algorithms have been proposed, e.g., \cite{ouyang2013stochastic,suzuki2013dual,wang2013online}. However, these algorithms all possess sublinear convergence rates.
Therefore, several recent works tried to combine the variance reduction scheme and ADMM to accelerate convergence.
For instance, SVRG-ADMM was proposed in \cite{zheng2016fast}. It was shown that   SVRG-ADMM converges linearly when the objective is strongly convex, and  has a sublinear convergence rate in the general convex case.
Another recent work \cite{zheng2016stochastic} further proved that SVRG-ADMM converges to a stationary point with a rate $O(\frac{1}{T})$ when the objective is nonconvex.
In \cite{qian2016accelerated}, the authors used acceleration technique in \cite{allen2016katyusha,hien2016accelerated} to further improve the convergence rate of SVRG-ADMM. However, all aforementioned variance-reduced ADMM algorithms cannot be directly applied to solving the stochastic composition optimization problem.

\subsection{Contribution}
In this paper, we propose an efficient algorithm called com-SVR-ADMM,  which combines ideas of SVRG and ADMM, to solve stochastic composition optimization. Our algorithm is based on the SVRG-ADMM algorithm proposed in \cite{zheng2016fast}, which does not apply to composite optimization problems. We consider three different objective functions in Problem (\ref{eq:problem-2-obj}), and show that our algorithm achieves the following performance.
\begin{itemize}
\item When $F(x)$ is strongly convex and Lipschitz smooth, and $R(\omega)$ is convex,  our algorithm converges linearly. This convergence rate is comparable with those of  com-SVRG-1 and com-SVRG-2  in \cite{lian2016finite}.
However, com-SVRG-1 and com-SVRG-2 do not take the commonly used regularization penalty into consideration. Experimental results also show that  com-SVR-ADMM  converges faster than com-SVRG-1 and com-SVRG-2.

\item   When $F(x)$ is convex and Lipschitz smooth, and $R(\omega)$ is convex,  com-SVR-ADMM has a sublinear rate of $O(\frac{\log(S+1)}{S})$, where $S$ is the number of outer iterations.\footnote{The number of inner iterations is constant.}
This result outperforms the $O(S^{-4/9})$ convergence rate of ASC-PG in \cite{wang2016accelerating}.\footnote{Note that ASC-PG is not based on SVRG and does not have inner loops.}

  \item When $F(x)$ and $R(\omega)$ are general convex functions (not necessarily Lipschitz smooth),  com-SVR-ADMM achieves a rate of $O(\frac{1}{\sqrt{S}})$. To the best of our knowledge, this is the first convergence result for stochastic composite optimization with general convex problems without  Lipschitz smoothness.
\end{itemize}

\subsection{Notation}
For vector $x$ and a positive definite matrix $G$, the $G$-norm of vector $x$ is defined as $||x||_{G} = \sqrt{x^TGx}$. For a matrix $X$, $||X||$ denotes its spectral norm, $\sigma_{max}(X), \sigma_{min}(X)$ denote its largest and smallest eigenvalue, respectively. $X^{\dag}$ denotes the pseudoinverse of $X$.
$\tilde{\nabla}R(\omega)$ denotes a noisy subgradient of nonsmooth $R(\omega)$. For a function $g(x): \mathbb{R}^{q} \mapsto \mathbb{R}^{r}$,  $\partial g(x) \in \mathbb{R}^{r\times q}$ denotes its Jacobian matrix. $\nabla f_{i_k}(g(x))$ denotes the gradient of $f_{i_k}(\cdot)$ at point $y = g(x)$.

\section{Algorithm}
\label{sec:alg}
Recall that the stochastic composition problem we want to solve has the following form:
\begin{eqnarray*}
&\min_{x,\omega}& F(x) + R(\omega) \\
&\text{s.t.}&  Ax+B\omega = 0.
\end{eqnarray*}
where $F(x)\triangleq\frac{1}{n} \sum\limits_{i=1}^{n} f_{i}(\frac{1}{m}\sum\limits_{j=1}^{m}g_j(x))$. For clarity, we denote $F(x) = \frac{1}{n}\sum\limits_{i=1}^nF_i(x)$, $F_i(x) = f_i(g(x))$, $g(x) = \mathbf{E}g(x) = \frac{1}{m}\sum\limits_{j=1}^m g_j(x)$.
Therefore, $\nabla F_i(x) = (\partial g(x))^T \nabla f_i(g(x))$.  

Our proposed procedure adopts the ADMM scheme. At every iteration the primal variables $(x,\omega)$ are obtained by minimizing the following augmented Lagrangian equation parameterized with  $\rho > 0$, i.e.,
\begin{eqnarray*}
  L_\rho(x,\omega,\lambda)
 = F(x) + R(\omega) + \langle\, \lambda, Ax+B\omega\rangle + \frac{\rho}{2}||Ax + B\omega||_2^2.
\end{eqnarray*}
The update of dual variable $\lambda$ is similar to that under gradient descent with the stepsize equaling $\rho$.
We also based our algorithm on a \emph{sampling oracle} as in \cite{wang2016accelerating}.  Specifically, we assume a stochastic first-order oracle, which, if queried,   returns a noisy gradient/subgradient or function value of $f_i(\cdot)$ and $g_j(\cdot)$ at a given point.

In the following sections, we introduce the stochastic variance reduced ADMM algorithm  for solving stochastic compositional optimization (com-SVR-ADMM). We present com-SVR-ADMM in three different scenarios, i.e., strongly convex and Lipschitz smooth, general convex and Lipschitz smooth, and general convex. Algorithm \ref{Algorithm-1} shows how com-SVR-ADMM is used in the strongly convex case, while Algorithm \ref{Algorithm-2} is for the second and third cases.

\subsection{Compositional Stochastic Variance Reduced ADMM for Strongly Convex Functions}
\begin{algorithm}[tb]
   \caption{com-SVR-ADMM for strongly convex stochastic composition optimization}
\begin{algorithmic}[1]   \label{Algorithm-1}
   \STATE {\bfseries Input:} $K$, $M$, $N$, $\eta$, $\rho$, $\tilde{x}^0$, $\tilde{\omega}^0$, $\tilde{\lambda}^0 = -(A^T)^{\dag}\nabla F(\tilde{x}^0)$;
   \FOR{$s=1,2,...$}
   \STATE $\tilde{x} = \tilde{x}^{s-1}$, \
   $x^0 = \tilde{x}^{s-1}$, \
   $\omega^0 = \tilde{\omega}^{s-1}$, \
   $\lambda^0 = \tilde{\lambda}^{s-1}$; \
   \STATE $g(\tilde{x}) = \frac{1}{m} \sum_{j=1}^{m}g_{j}(\tilde{x})$; \quad \quad \quad \quad  ($m$ queries)
   \STATE evaluate $\nabla F(\tilde{x})$;  \quad \quad \quad \quad \quad \quad \ ($m+n$ queries)
   \FOR{$k=0$ {\bfseries to} $K-1$}

   \STATE $\omega^{k+1} = \arg\min_{\omega} R(\omega) + \langle\,\lambda^{k}, B\omega\rangle + \frac{\rho}{2}||Ax^{k}+ B\omega||_2^2$;
   \medskip

   \STATE uniformly sample $N_k$ and calculate $\hat{g}(x^{k})$ using (\ref{eq:g_hat}); \qquad ($2N$ queries)
   \medskip
   \STATE uniformly sample $i_k$, $j_k$ and calculate $\nabla\hat{F}_{i_k}(x^k)$ using (\ref{eq:gra_f_a1});  \qquad ($4$ queries)
   \medskip
   \STATE $x^{k+1} = \arg\min_{x} \langle\,\nabla\hat{F}_{i_k}(x^k), x-x^k \rangle + \langle\,\lambda^{k}, Ax\rangle + \frac{\rho}{2}||Ax+B\omega^{k+1}||_2^2 + \frac{1}{2\eta}||x-x^k||_2^2$;
   \medskip

   \STATE $\lambda^{k+1} = \lambda^{k} + \rho(Ax^{k+1} + B\omega^{k+1})$;
   \ENDFOR
   \STATE $\tilde{x}^{s} = \frac{1}{K} \sum\limits_{k=1}^{K} x^{k}$, \
          $\tilde{w}^{s} = \frac{1}{K} \sum\limits_{k=1}^{K} w^{k}$, \
          $\tilde{\lambda}^{s} = -(A^T)^{\dag}\nabla F(\tilde{x}^s)$;
   \ENDFOR
   \STATE {\bfseries Output:} $\tilde{x}^{s}$, \ $\tilde{w}^{s}$.
\end{algorithmic}
\end{algorithm}
\noindent
\noindent
As in SVRG, com-SVR-ADMM has $K$ inner loops inside each outer iteration. At every outer iteration, we need to keep track of a reference point $\tilde{x}$ (Step $3$ in Algorithm~\ref{Algorithm-1}) for computing $g(\tilde{x})$ defined as
\begin{equation}
g(\tilde{x}) = \frac{1}{m}\sum\limits_{j=1}^{m} g_{j}( \tilde{x} ),
\end{equation}
and evaluate $\partial g(\tilde{x})$, which is the Jacobian matrix of $g(x)$ at point $\tilde{x}$. The updates of $\omega^{k+1}$ and $\lambda^{k+1}$ are the same as those in batch ADMM \cite{boyd2011distributed}.
The main difference lies in the update for $x^{k+1}$, in that   we use a stochastic sample $i_k$ and replace $F_{i_k}(x)$ with its first-order approximation, and then approximate $\nabla F_{i_k}(x^k)$ by
\begin{eqnarray}
  \label{eq:gra_f_a1}
 \nabla\hat{F}_{i_k}(x^k) =  (\partial g_{j_k}(x^k))^{T}\nabla f_{i_k}(\hat{g}(x^k))\label{eq:nabla_F_hat}
 -(\partial g_{j_k}(\tilde{x}) )^{T}\nabla f_{i_k}(g(\tilde{x})) +\nabla F(\tilde{x}).
\end{eqnarray}
Here $i_k$, $j_k$ are uniformly sampled from $\{ 1,2,...,n \}$ and $\{1,...,m\}$, respectively. $\hat{g}(x^k)$ is an estimation of $g(x^{k})$ defined as follows:
\begin{equation}
\hat{g}(x^{k}) = g(\tilde{x}) - \frac{1}{N} \sum\limits_{1\leq j \leq N} \left( g_{\mathbb{N}_k[j]}(\tilde{x}) -  g_{\mathbb{N}_k[j]}(x^{k})\right),
\label{eq:g_hat}
\end{equation}
where $\mathbb{N}_k$ is a mini-batch and is obtained by uniformly and randomly sampling from $\{1,...,m\}$ for $N$ times (with replacement) and $\mathbb{N}_k[j]$ is the $j$th element of $\mathbb{N}_k$.
In step $10$ of Algorithm \ref{Algorithm-1}, we add a proximal term $\frac{1}{2\eta}||x-x^k||_2^2$ to control the distance between the next point $x^{k+1}$ and the current point $x^k$. The parameter $\eta$ is a constant and plays the role of stepsize as in  \cite{ouyang2013stochastic}.

Note that our estimated gradient $\nabla\hat{F}_{i_k}(x^k)$  is biased due to the composition objective function and its form is the same as com-SVRG-1 in \cite{lian2016finite}. However, we remark that our algorithm is not a trivial extension of com-SVRG-1 due to the existence of linear constraint and Lagrangian dual variable. Moreover, com-SVR-ADMM can handle regularization penalty while com-SVRG-1 cannot. Also,  the update of $\tilde{\lambda}^s$ uses the pseudoinverse of $A$. In the common case when $A$ is sparse, one can use the efficient Lanczos algorithm \cite{golub2012matrix} to compute $A^{\dag}$.
Note that step $10$ in Algorithm~\ref{Algorithm-1} often involves computing $A^TA$. The memory complexity for this step can be alleviated by algorithms proposed in recent works, e.g., \cite{zheng2016fast,zhang2011unified}.
%

\subsection{Compositional Stochastic Variance Reduced ADMM  for General Convex Functions}
\label{sec:2.2}
In this section, we describe how com-SVR-ADMM handles general convex composition problems with Lipschitz smoothness. Without strong convexity, we need to make subtle changes.
As shown in Algorithm~\ref{Algorithm-2}, besides changes in variable initialization and output, another key difference is the approximation of $\nabla F_{i_k}(x)$, where we use $g(x^k)$ instead of $\hat{g}(x^k)$, i.e.,
\begin{eqnarray}
  \label{eq:gra_F_a2}
 \nabla\hat{F}_{i_k}(x^k) =  (\partial g_{j_k}(x^k))^T \nabla f_{i_k}(g(x^k))
- (\partial g_{j_k}(\tilde{x}))^T \nabla f_{i_k}(g(\tilde{x}))+ \nabla F(\tilde{x}).
\end{eqnarray}
Note that in the cases of interest (see Assumption \ref{assump:1} below), the approximated gradient $\nabla\hat{F}_{i_k}(x^k)$ is unbiased.
The next change is the stepsize for updating $x$. In step $10$ of Algorithm~\ref{Algorithm-2}, we use a positive definite matrix $G_k$ in the proximal term.\footnote{The corresponding proximal term of Algorithm~\ref{Algorithm-1} can be viewed to have $G_k = I$.} Therefore, the stepsize depends on two parameters: $\eta_s$ and $G_k$, as shown in (\ref{eq:eta_G_1}), where $s$ and $k$ are the iteration counters for outer and inner iteration, respectively.
Here $L_F$ is a parameter of Lipschitz smoothness and will be  specified in our assumptions in next section.
\begin{equation}
\begin{aligned}
  \label{eq:eta_G_1}
\eta_s = \frac{1}{(s+1)L_F},\quad &G_0 \succeq G_1 \succeq  G_2 \succeq ... \succeq G_{K-1},\\
G_0 = \frac{1}{s}I,\quad G_{K-1} &= \frac{1}{s+1}I,\quad  G_{K} = \frac{1}{s+1}I.
\end{aligned}
\end{equation}
That is, $G_k$ is nonincreasing for $k = 0,1,...,K$. Then, according to the definition of $G$-norm and (\ref{eq:eta_G_1}), we have:
\begin{equation}
  \label{eq:stepsize-2}
  \frac{1}{2\eta_s}||x-x^k||^2_{G_k} = \frac{1} {2\eta_{s,k}} ||x-x^k||^2_2,
\end{equation}
where $\eta_{s,k} = \frac{\eta_s}{{\bf co}(G_k)}$ and ${\bf co}(G_k) = a$ if $G_k = aI$, and $a$ is a scalar. Therefore, $\eta_{s,k}$ serves as the stepsize \cite{ouyang2013stochastic}, and it can be verified that $\eta_{s,k}$ satisfies the following properties:
\begin{equation}
  \begin{aligned}
&&  \eta_{s,0} = \frac{s}{(s+1)L_F},\ \eta_{s,K-1} &= \frac{1}{L_F}, \ \eta_{s,K} = \frac{1}{L_F},\\
&&  \eta_{s,0} \leq \eta_{s,1} \leq ... \leq &\eta_{s,K-1}.
\end{aligned}
\end{equation}
That is, $\eta_{s,k}$ changes from $\frac{s}{(s+1)L_F}$ to $\frac{1}{L_F}$ in stage $s$.
Note that even though $\eta_{s,k}$ is not a constant, it still has a reasonable value and does not need to vanish. This feature is helpful for convergence acceleration.

\begin{algorithm}[tb]
   \caption{com-SVR-ADMM for general convex stochastic composition optimization}
   \label{Algorithm-2}
\begin{algorithmic}[1]
   \STATE {\bfseries Input:} $S$, $K$, $N$, $\eta_s$, $\rho$, $\tilde{x}^0 = \hat{x}^0$, $\hat{\omega}^0$, $\hat{\lambda}^0$, $\hat{G}^0 = I$;
   \FOR{$s=1,2,...,S$}
   \STATE $\tilde{x} = \tilde{x}^{s-1}$, \
   $x^0 = \hat{x}^{s-1}$, \
   $\omega^0 = \hat{\omega}^{s-1}$, \
   $\lambda^0 = \hat{\lambda}^{s-1}$, \
   $G_0 = \hat{G}^{s-1}$;\

   \STATE $g(\tilde{x}) = \frac{1}{m} \sum_{j=1}^{m}g_{j}(\tilde{x})$; \qquad \qquad \qquad ($m$  queries)
   \STATE evaluate $\nabla F(\tilde{x})$; \qquad \qquad \qquad \qquad \ ($m+n$ queries)
   \FOR{$k=0$ {\bfseries to} $K-1$}
   \STATE $\omega^{k+1} = \arg\min_{\omega} R(\omega) + \langle\,\lambda^{k}, B\omega\rangle + \frac{\rho}{2}||Ax^{k}+ B\omega||_2^2$;
   \medskip
   \STATE calculate $g(x^{k}) = \frac{1}{m} \sum_{j=1}^{m}g_{j}(x^k)$;   \qquad ($m$ queries)
   \STATE uniformly sample $i_k$, $j_k$ and calculate $\nabla\hat{F}_{i_k}(x^k)$ using (\ref{eq:gra_F_a2}); \qquad  ($4$ queries)
   \medskip
   \STATE $x^{k+1} = \arg\min_{x} \langle\,\nabla\hat{F}_{i_k}(x^k), x-x^k \rangle + \langle\,\lambda^{k}, Ax\rangle + \frac{\rho}{2}||Ax+B\omega^{k+1}||_2^2 + \frac{1}{2\eta_s}||x-x^k||_{G_k}^2$;
   \\
   \STATE $\lambda^{k+1} = \lambda^{k} + \rho(Ax^{k+1} + B\omega^{k+1})$;
   \ENDFOR
   \STATE $\tilde{x}^{s} = \frac{1}{K} \sum\limits_{k=1}^{K} x^{k}$, \
          $\tilde{\omega}^{s} = \frac{1}{K} \sum\limits_{k=1}^{K} \omega^{k}$, \
          $\tilde{\lambda}^{s} = \frac{1}{K} \sum\limits_{k=1}^{K} \lambda^{k}$,
          $\hat{x}^s = x^K,\ \hat{\omega}^s = \omega^K,\ \hat{\lambda}^s = \lambda^K$, \ $\hat{G}^s = G_K$;
   \ENDFOR
   \STATE {\bfseries Output:} $\bar{x} = \frac{1}{S}\sum\limits_{s=1}^{S}\tilde{x}^s$, \
                              $\bar{\omega} = \frac{1}{S}\sum\limits_{s=1}^{S}\tilde{\omega}^s$.
\end{algorithmic}
\end{algorithm}
\noindent
\subsection{General Convex Functions without Lipschitz Smoothness}
In the previous two sections, we present the procedures of com-SVR-ADMM for the strongly convex and general convex settings, both under the Lipschitz smooth assumption of $F(x)$. In this section, we further investigate the case when the smooth condition is relaxed. We still use Algorithm~\ref{Algorithm-2},  except that the values $\eta_s$ and $G_k$ are changed to
\begin{equation}
\begin{aligned}
  \label{eq:eta_G_2}
\eta_s = \frac{1}{s+1},\quad &G_0 \succeq G_1 \succeq  G_2 \succeq ... \succeq G_{K-1},\\
G_0 = \frac{1}{\sqrt{s}}I,\quad  G_{K-1} &= \frac{1}{\sqrt{s+1}}I,\quad  G_{K} = \frac{1}{\sqrt{s+1}}I.
\end{aligned}
\end{equation}
 Therefore, using the same technique in (\ref{eq:stepsize-2}), it can be verified that $\eta_{s,k}$ in this setting changes from $\frac{\sqrt{s}}{s+1}$ to $\frac{1}{\sqrt{s+1}}$ in stage $s$ and decreases to zero.
Note that in this case, the number of oracle calls at each step  is the same as that in section \ref{sec:2.2}.

Although the algorithm we proposed appears similar to the SVRG-ADMM algorithm in \cite{zheng2016fast},   it is very different due to the composition nature of the objective function (which is not considered in SVRG-ADMM) and the stochastic variance reduced gradients in (\ref{eq:gra_f_a1}) and (\ref{eq:gra_F_a2}).
These differences make it impossible to directly apply SVRG-ADMM and require a very different analysis for the new algorithm. Readers interested in the full  proofs can refer to the appendix.

\section{Theoretical Results}
In this section, we analyze the  convergence performance of com-SVR-ADMM under the three cases described in section \ref{sec:alg}. 
Below, we first state our assumptions. Note that the assumptions are not restrictive and are commonly made in the literature, e.g., \cite{wang2016accelerating,ouyang2013stochastic,wang2017stochastic,zheng2016stochastic}.

\begin{assumption}
  \label{assump:1}
(i) For each $i \in \{1,...,n\}$, $F_i$ is convex and continuously differentiable, $R(\omega)$ is convex (can be nonsmooth). Moreover, there exists an optimal primal-dual solution $(x^*, \omega^*,\lambda^*)$ for Problem (\ref{eq:problem-2-obj}).

(ii) The feasible set $\mathbb{X}$ for $x$  is bounded and denote $D = \max_{x,y \in \mathbb{X}}||x-y||$.

(iii) For randomly sampled $i_k \in \{1,...,n\}$, $j_k \in \{1,...,m\}$ and $\forall x$, we assume the following unbiased properties:
\begin{equation}
 \begin{aligned}
 E((\partial g_{j_k}(x))^T\nabla f_{i_k}(g(x))) &= \nabla F(x),\\
 E(\partial g_{j_k}(x)) = \partial g(x), \quad E(\nabla &F_{i_k}(x)) = \nabla F(x).
\end{aligned}
 \end{equation}
\end{assumption}

\begin{assumption}
  \label{assump:2}
$F$ is strongly convex with parameter $\mu_F > 0$, i.e., $\forall x$,
\begin{equation}
F(x) - F(x^*) \geq \langle\, \nabla F(x^*), x-x^*\rangle + \frac{\mu_F}{2}||x-x^*||_2^2.
\end{equation}
\end{assumption}
\begin{assumption} \label{assump:3} Matrix $A$ has full row rank.
\end{assumption}

\begin{assumption} \label{assump:7} There exists a positive constant $L_F$, such that $\forall i \in \{1,...,n\}$, $\forall j \in \{1,...,m\}$ and $\forall x$, $y$, we have
 \begin{equation}
||(\partial g_j(x))^T\nabla f_i(g(x)) - (\partial g_j(y))^T\nabla f_i(g(y))|| \\
\leq L_F||x-y||.\nonumber
\end{equation} 
\end{assumption}

\begin{assumption} \label{assump:4} For each $i \in \{1,...,n\}$, $f_i$ is Lipschitz smooth with positive parameter $L_f$, that is, $\forall x,y$, we have
\begin{equation}
  ||\nabla f_{i}(y) - \nabla f_{i}(x)|| \leq L_f||y-x||.
\end{equation}
\end{assumption}

\begin{assumption}
 \label{assump:5}
For every $j \in \{1,...,m\}$, $\partial g_j(x)$ is bounded, and for all $ x,y$, $\exists \ C_G, L_G > 0$ that satisfy
 \begin{equation}
   \begin{aligned}
   ||g_{j}(x) - g_{j}(y)|| \leq C_G||x-y||, \quad ||\partial g_{j}(x)||\leq C_G, \\
     ||g_{j}(x) - g_{j}(y)|| \leq L_G||x-y||^2. \qquad \qquad
   \end{aligned}
 \end{equation}
\end{assumption}
\indent For clarity, we also use the following  notations used in the   theorems:
\begin{equation}
\begin{aligned}
u = \left[ \begin{matrix}  x \\ \omega   \end{matrix} \right], \
u^k = \left[ \begin{matrix}  x^k \\ \omega^k   \end{matrix} \right], \
&\tilde{u}^s = \left[ \begin{matrix}  \tilde{x}^s \\ \tilde{\omega}^s   \end{matrix} \right], \
\bar{u} = \left[ \begin{matrix}  \bar{x} \\ \bar{\omega}   \end{matrix} \right],
\\
G(u) =  F(x) - F(x^*) - \langle\, \nabla F(x^*), x-x^*\rangle\
 + \ & R(\omega) - R(\omega^*) - \langle\, \tilde{\nabla}R(\omega^*), \omega-\omega^*\rangle.
\end{aligned}
\end{equation}

It can be verified that $G(u)$ is always non-negative due to the convexity of $F(x)$ and $R(\omega)$.  The following theorem and corollary show that Algorithm \ref{Algorithm-1} has a linear convergence rate.
\begin{proposition*}
Under Assumption \ref{assump:7}, we have  $\forall i \in  \{1,...,n\}$ and $\forall x,y$:
\begin{equation}
 \label{eq:F_i_lip}
||\nabla F_i(x) - \nabla F_i(y)|| \leq L_F||x-y||,
\end{equation}
i.e., each $F_i$ is Lipschitz smooth. Moreover, it   implies $||\nabla F(x) - \nabla F(y)|| \leq L_F||x-y||$.
\end{proposition*}

\begin{theorem}
  \label{thm:1}
Under Assumptions \ref{assump:1}, \ref{assump:2}, \ref{assump:3}, \ref{assump:7}, \ref{assump:4} and \ref{assump:5},  if\ $0 < \eta \leq 1/L_F$, then under Algorithm~\ref{Algorithm-1},
\begin{equation}
 \gamma_1\mathbf{E}[G(\tilde{u}^s)] \leq \gamma_2G(\tilde{u}^{s-1}),
\end{equation}
where (denote $\sigma(N) = \sqrt{1/N}$)
\begin{equation*}
\begin{aligned}
\gamma_1 = &(2\eta - \frac{32\eta^2C_G^4L_f^2}{\mu_F N} -\frac{48\eta^2L_F^2+8\eta DC_GL_fL_G\sigma(N)}{\mu_F})K,\\
\gamma_2 = &(K+1)(\frac{32\eta^2C_G^4L_f^2}{\mu_F N} + \frac{48\eta^2L_F^2+8\eta DC_GL_fL_G\sigma(N)}{\mu_F})\\
&+\frac{2}{\mu_F} + \frac{2\eta \rho ||A^TA||}{\mu_F} + \frac{2L_F\eta}{\rho\sigma_{min}(AA^T)}.
\end{aligned}
\end{equation*}
 \end{theorem}
 \begin{corollary}\label{corollary:linear-conv}
Suppose the conditions in  Theorem~\ref{thm:1} hold. Then, there exist  positive $\Theta(1)$ constants $K$ (number of inner iterations) and $N$ (mini-batch size) such that $\gamma_1, \gamma_2 >0, \gamma = \gamma_2 / \gamma_1  < 1$. Thus,  Algorithm~\ref{Algorithm-1} converges linearly.
\end{corollary}
From Corollary \ref{corollary:linear-conv}, if we want to achieve $\mathbf{E}[G(\tilde{u}^s) ] \leq \epsilon$, $\forall \epsilon > 0$, the number of steps we need to take is roughly $s \geq \log(\frac{G(\tilde{u}^0)}{\epsilon})/\log(\frac{1}{\gamma})$. In each iteration, we need $2m+n + K(2N + 4)$ oracle calls.
Therefore, the overall query complexity is $O((m+n+KN)\log{\frac{1}{\epsilon}})$.
For comparison,   the query complexity is $O((m+n+\kappa^4)\log(1/\epsilon))$ for com-SVRG-1 and $O((m+n+\kappa^3)\log(1/\epsilon))$ for com-SVRG-2 \cite{lian2016finite}, where $\kappa$ is a parameter related to condition number.
We will see in simulations in section \ref{section:simulation} that the overall query complexity of com-SVR-ADMM is lower than com-SVRG-1 and com-SVRG-2. 

Now we prove the convergence property of com-SVR-ADMM under Assumptions \ref{assump:1} and \ref{assump:7}.

\begin{theorem}
\label{thm:2}\ Under Assumptions \ref{assump:1} and \ref{assump:7} ,  if  $\eta_s$ and $G_k$ are chosen as  in (\ref{eq:eta_G_1}), under Algorithm \ref{Algorithm-2},
\begin{equation}
\begin{aligned}
  \mathbf{E}&(G(\bar{u}) + \Lambda||A\bar{x} + B\bar{\omega}||)\\
  \leq &\frac{4L_FD^2\log(S+1)}{S} + \frac{L_FD^2\log{S}}{2KS} +\frac{L_FD^2 + \rho D^2||A^TA|| + \frac{2}{\rho} ||\hat{\lambda}^0 - \lambda^*||_2^2 + \frac{2}{\rho}\Lambda^2}{2KS},
\end{aligned}
\end{equation}
where $\Lambda > 0$.
\end{theorem}

From Theorem \ref{thm:2}, we see that com-SVR-ADMM has an $O(\frac{\log(S+1)}{S})$ convergence rate under the general convex and Lipschitz smooth condition. It improves upon the convergence rate $O(S^{-4/9})$ in the recent work \cite{wang2016accelerating}.
In Theorem \ref{thm:2}, we consider both the convergence property of function value and feasibility violation. Since $G(u)$ and $||A\bar{x}+B\bar{\omega}||$ are both non-negative, each term has an $O(\frac{\log(S+1)}{S})$ convergence rate.

In the following theorem, we show that our algorithm exhibits $O(\frac{1}{\sqrt{S}})$ convergence rate for both the objective value and feasibility violation, when the objective  is a general  convex function.

\begin{assumption}
 \label{assump:8}
The gradients/subgradients of all $f_i$, $F_i$, $g_j$ and $R(\omega)$ are bounded and $||\nabla F_i(x)|| \leq C_F$, \ $C_F > 0$.
Moreover, $B$ is invertible and $A,B$ are bounded.
\end{assumption}
\begin{theorem}
  \label{thm:3}
Under Assumptions \ref{assump:1} and \ref{assump:8}, denote\\
$\dot{x}^s = \frac{1}{K}\sum\limits_{k=0}^{K-1}x^k$,
$\tilde{z}^s = \left[ \begin{matrix}  \dot{x}^s \\ \tilde{w}^s  \end{matrix} \right]$,
$\bar{z} = \frac{1}{S}\sum\limits_{s=1}^{S}\tilde{z}^s$.
If $\eta_s$ and $G_k$ are chosen as in (\ref{eq:eta_G_2}), there exists a positive $\Theta(1)$ constant $\rho$ such that, under Algorithm \ref{Algorithm-2},
\begin{equation}
\begin{aligned}
 \mathbf{E}&(G(\bar{z}) + \Lambda||A\bar{x} + B\bar{\omega}||)\\
  \leq &\frac{C_1(C_4+C_F)}{\sqrt{S}} + \frac{D^2}{K\sqrt{S}}  + \frac{ C_3\log(S+1) }{S} + \frac{D^2+\rho||A^TA||D^2 + \frac{2}{\rho}||\hat{\lambda}^0 - \lambda^*||_2^2 +\frac{2}{\rho}\Lambda^2}{2KS},
\end{aligned}
\end{equation}
where $\Lambda$, $C_1$, $C_3$, $C_4$ are positive constants.
\end{theorem}
%
The reason for the introduction of $\bar{z}$  is similar to the step taken in \cite{ouyang2013stochastic}, and is due to the lack of Lipschitz smooth property. This result  implies an $O(\frac{1}{\sqrt{S}})$ convergence rate for both objective value and feasibility violation.

\section{Experiments}
\label{section:simulation}
In this section, we conduct experiments and compare com-SVR-ADMM to existing algorithms. We consider two experiment scenarios, i.e., the portfolio management scenario from \cite{lian2016finite} and the reinforcement learning scenario from \cite{wang2016accelerating}. Since the objective functions in both scenarios are strongly convex and Lipshitz smooth, we only provide results for Algorithm \ref{Algorithm-1}.

\subsection{Portfolio Management}
Portfolio management is usually formulated as mean-variance minimization of the following form:
\begin{equation}
  \min_{x} -\frac{1}{n} \sum\limits_{i=1}^n\langle\, r_i,x\rangle + \frac{1}{n}\sum\limits_{i=1}^{n}(\langle\, r_i,x\rangle - \frac{1}{n}\sum\limits_{j=1}^{n}\langle\, r_j,x\rangle)^2 + R(x),
\end{equation}
where $r_i \in \mathbb{R}^N$ for $i \in \{ 1,...,n\}$, $N$ is the number of assets, and $n$ is the number of observed time slots. Thus,  $r_i$ is the observed reward in time slot $i$.
We compare our proposed com-SVR-ADMM with three benchmarks: com-SVRG-1, com-SVRG-2 from \cite{lian2016finite}, and SGD. In order to compute the unbiased stochastic gradient of SGD, we first enumerate all samples in the data set of $g$ to calculate $g(x)$ and $\partial g(x)$, then evaluate $\partial g(x)\nabla f_i(g(x))$ for a random sample $i$. Using the same definition of $g_j(x)$ and $f_i(y)$ and the same parameters generation method as \cite{lian2016finite}, we  set the regularization to $R(x) = \frac{\mu}{2}||x||_2^2$, where $\mu > 0$.

\begin{figure}[!tb]
\vskip 0.2in
\minipage{0.46\textwidth}
  \includegraphics[width=\linewidth]{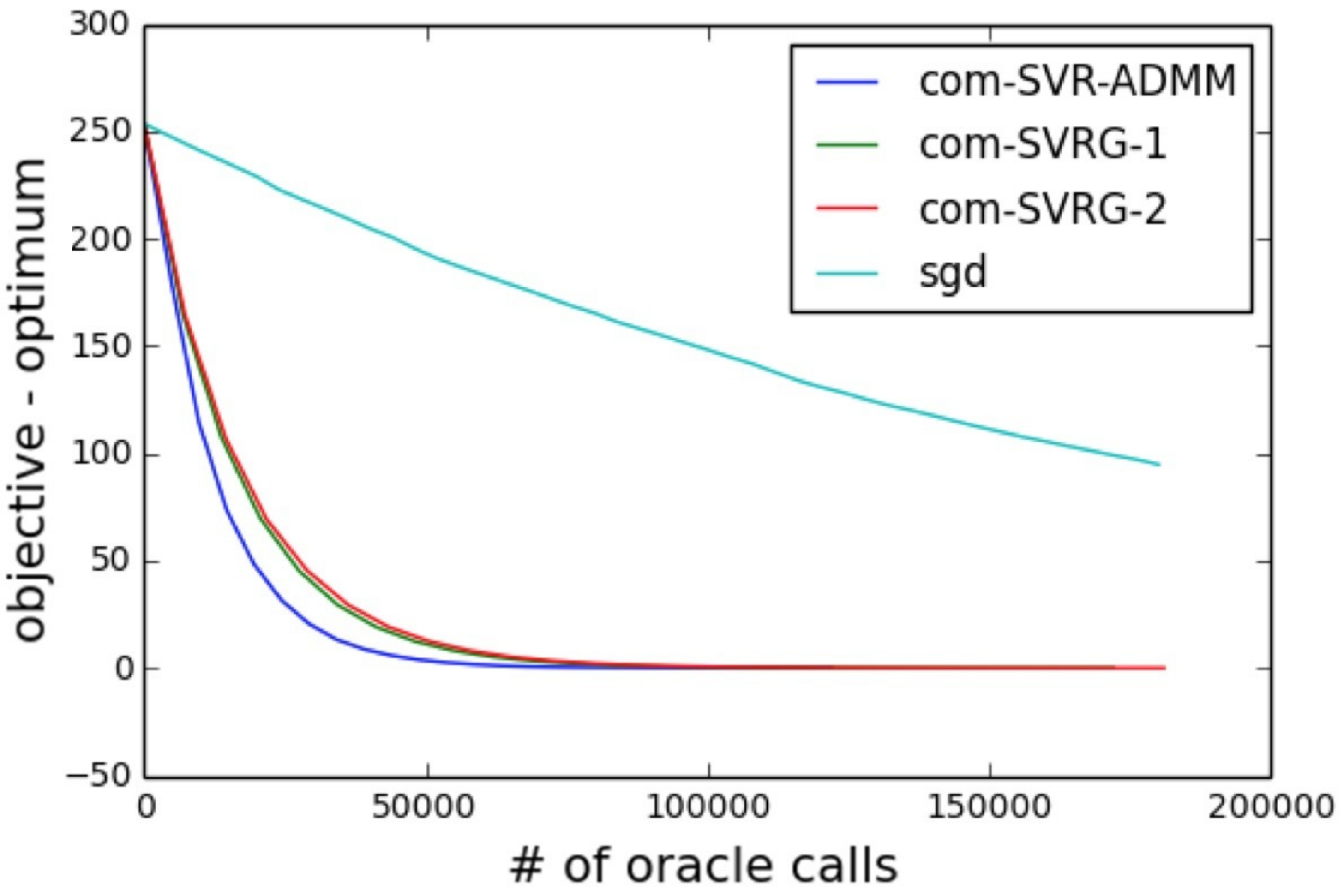}
  \label{fig:pt-1-1}
\endminipage\hfill
\minipage{0.46\textwidth}
  \includegraphics[width=\linewidth]{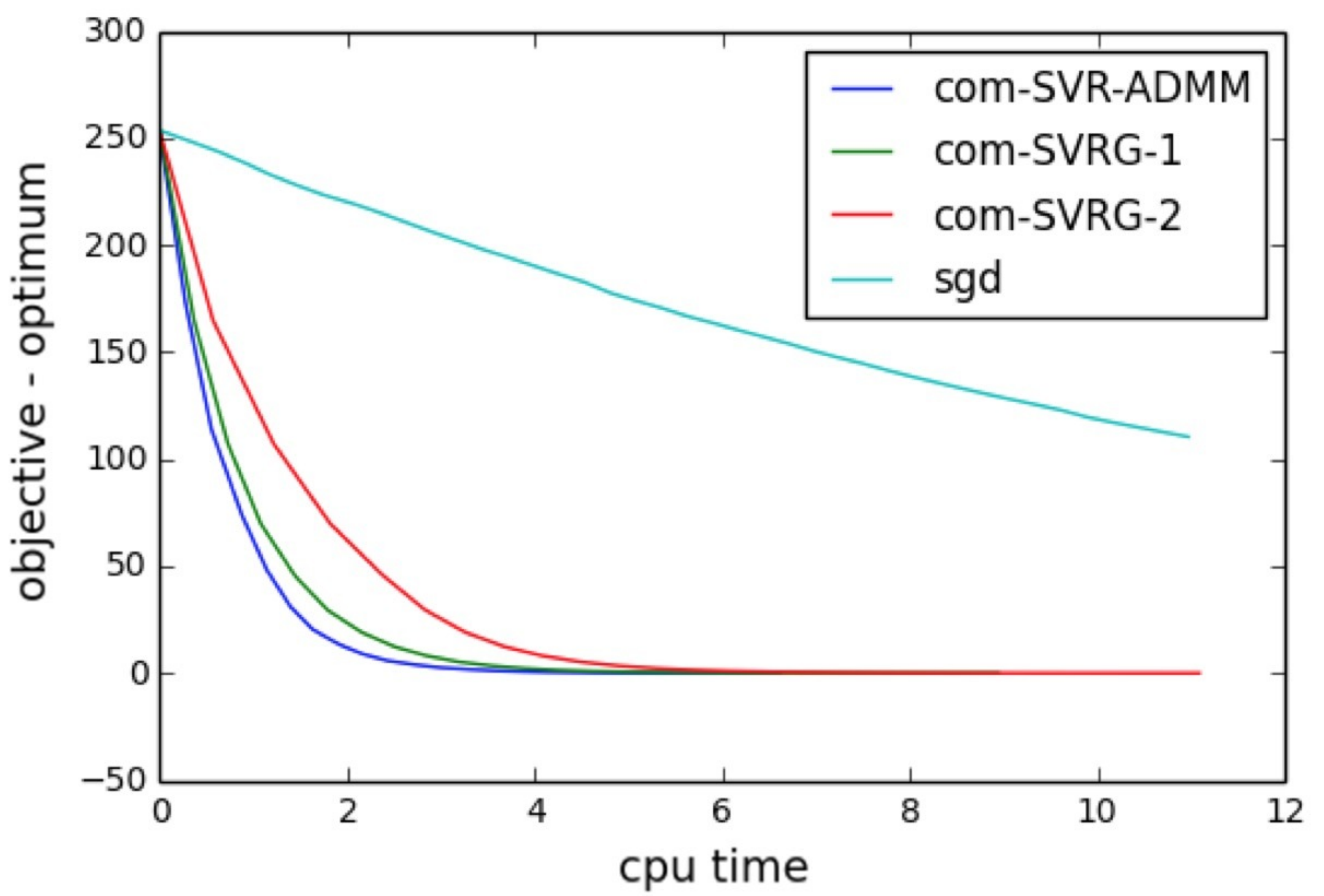}
  \label{fig:pt-1-2}
\endminipage
\caption{Portfolio Management with $cov=2$.}
\vskip -0.2in
\label{pt-1}
\end{figure}

\begin{figure}[!tb]
\vskip 0.2in
\minipage{0.46\textwidth}
  \includegraphics[width=\linewidth]{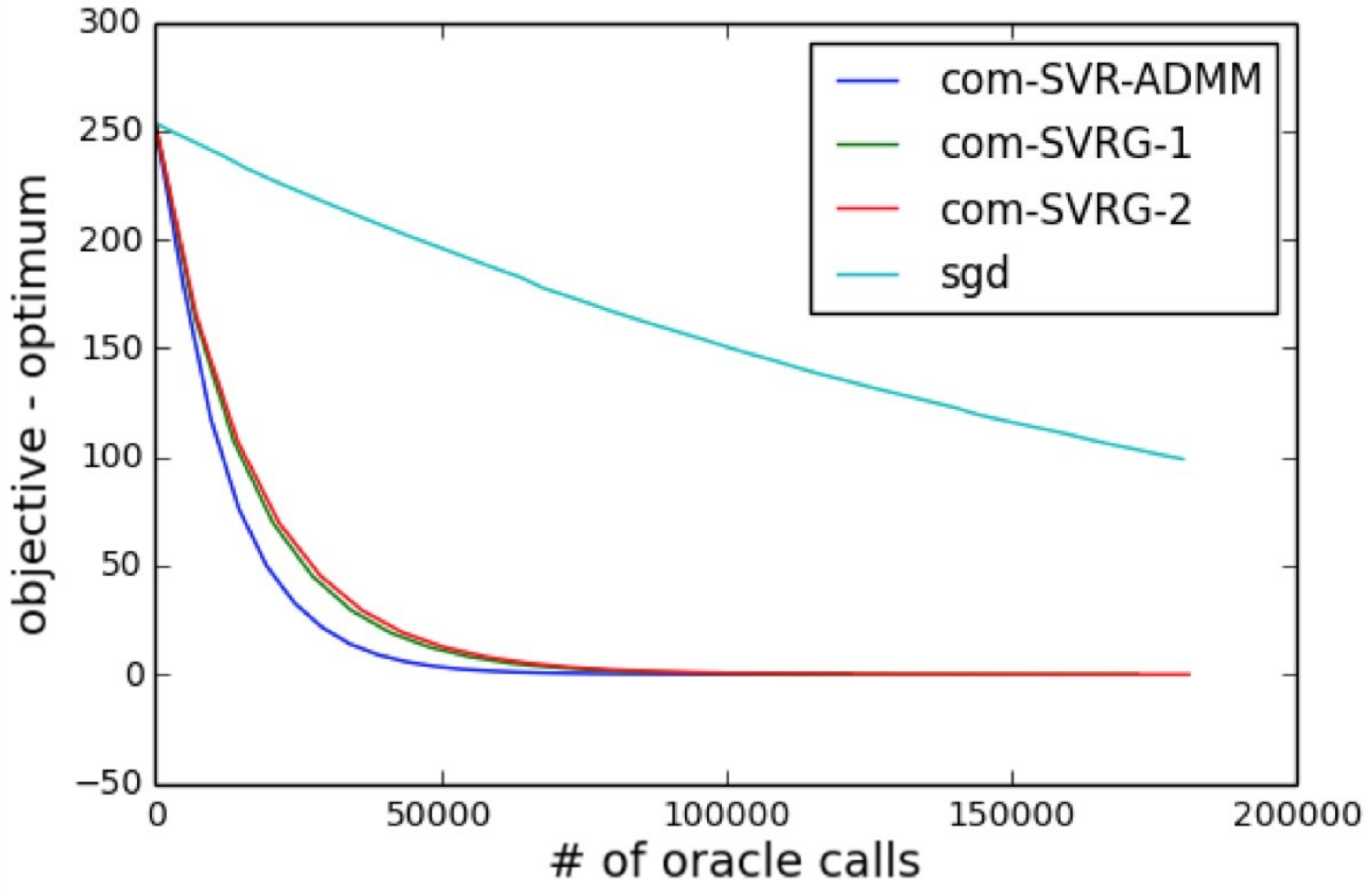}
  \label{fig:pt-2-1}
\endminipage\hfill
\minipage{0.46\textwidth}
  \includegraphics[width=\linewidth]{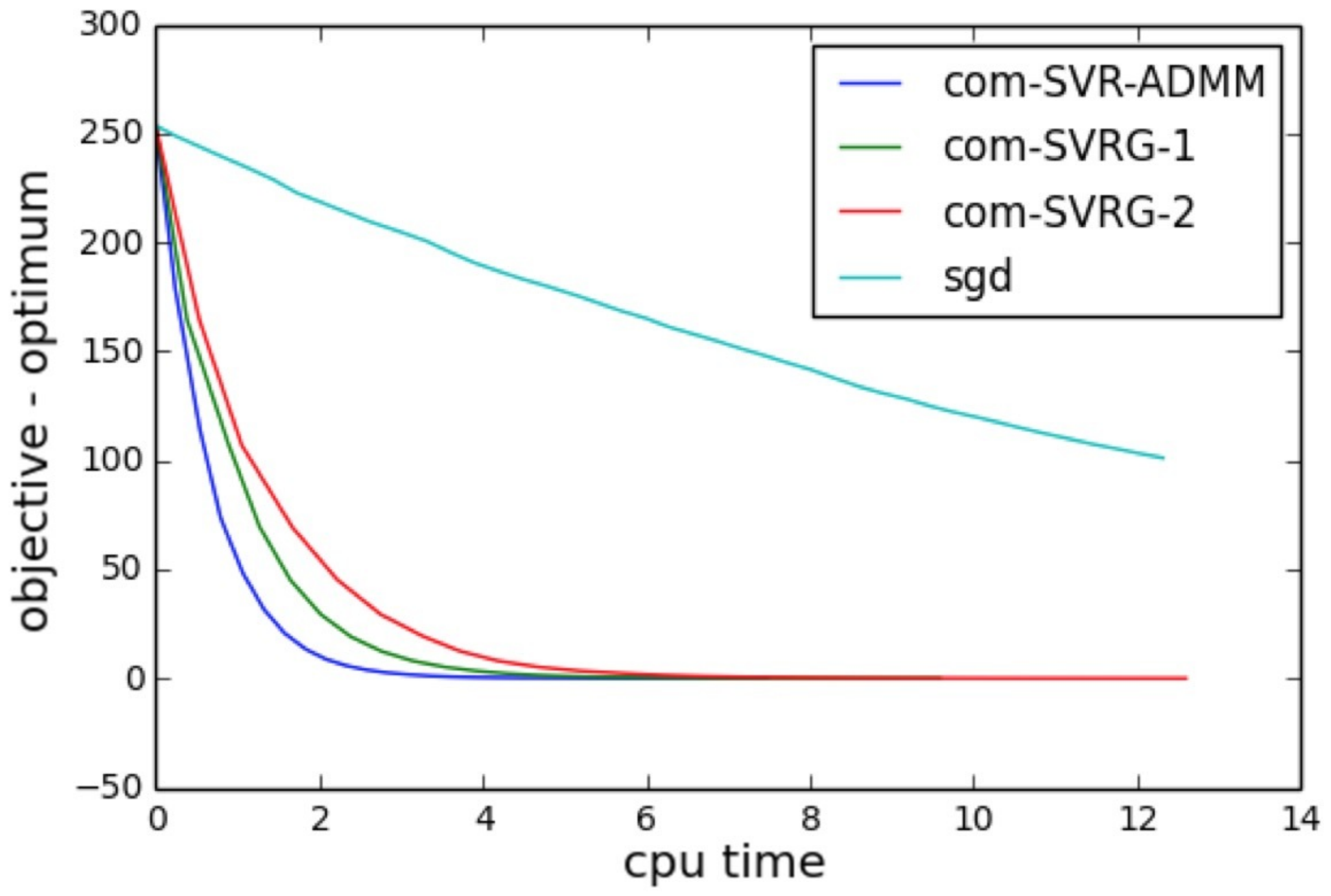}
  \label{fig:pt-2-2}
\endminipage
\caption{Portfolio Management with $cov=10$. The other parameters have the same value as Figure~\ref{pt-1}.}
\vskip -0.2in
\label{pt-2}
\end{figure}

The experimental results are shown in Figure~\ref{pt-1} and Figure~\ref{pt-2}. Here the $y$-axis represents the objective value minus optimal value and the $x$-axis is the number of oracle calls or CPU time.
We set $N=200$, $n=2000$.   \emph{cov} is the parameter used for reward covariance matrix generation \cite{lian2016finite}. In Figure~\ref{pt-1}, $cov=2$, and $cov=10$ in Figure~\ref{pt-2}. All shared parameters in the four algorithms, e.g., stepsize, have the same values.
We can see that all SVRG based algorithms perform much better than SGD,  and  com-SVR-ADMM outperforms   two other linear convergent algorithms.
\noindent
\noindent
\begin{figure}[!tb]
\vskip 0.2in
\minipage{0.46\textwidth}
  \includegraphics[width=\linewidth]{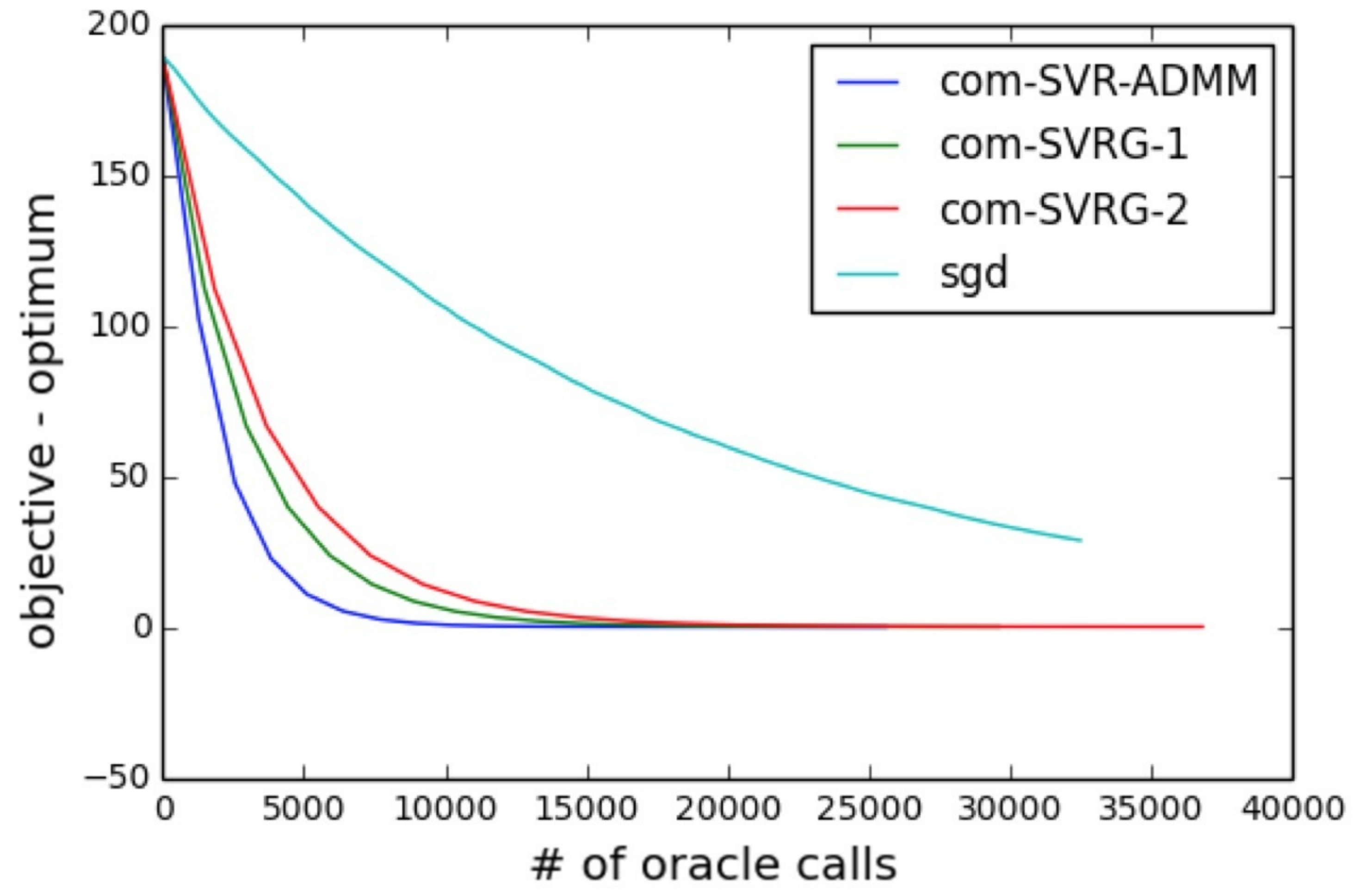}
  \label{fig:2-1}
\endminipage\hfill
\minipage{0.46\textwidth}
  \includegraphics[width=\linewidth]{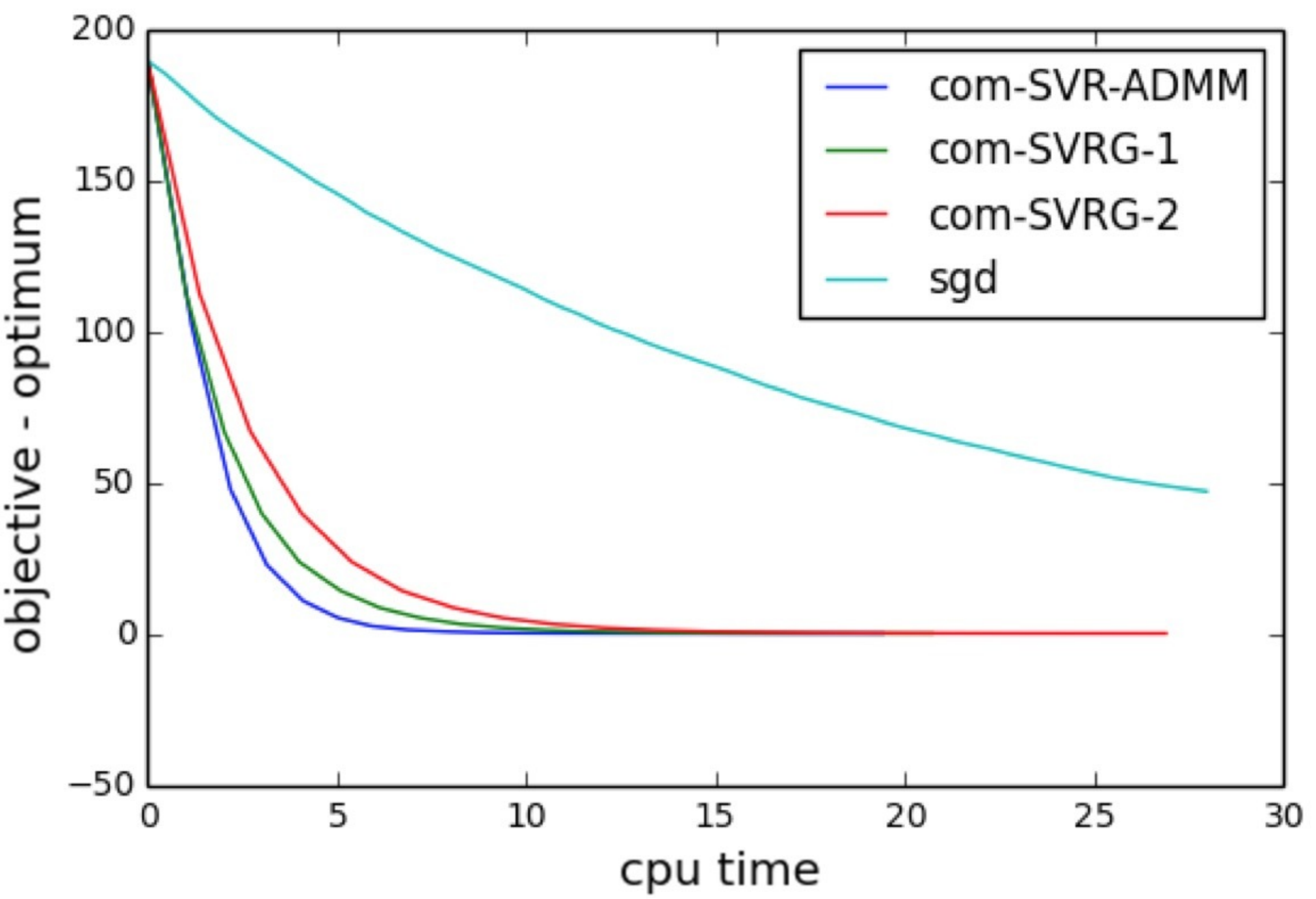}
  \label{fig:rl-1-2}
\endminipage
\caption{On-policy learning experiment with $S=200$, $d=100$.}
\vskip -0.1in
\label{rl-1}
\end{figure}
\noindent
\begin{figure}[!tb]
\vskip 0.2in
\minipage{0.46\textwidth}
  \includegraphics[width=\linewidth]{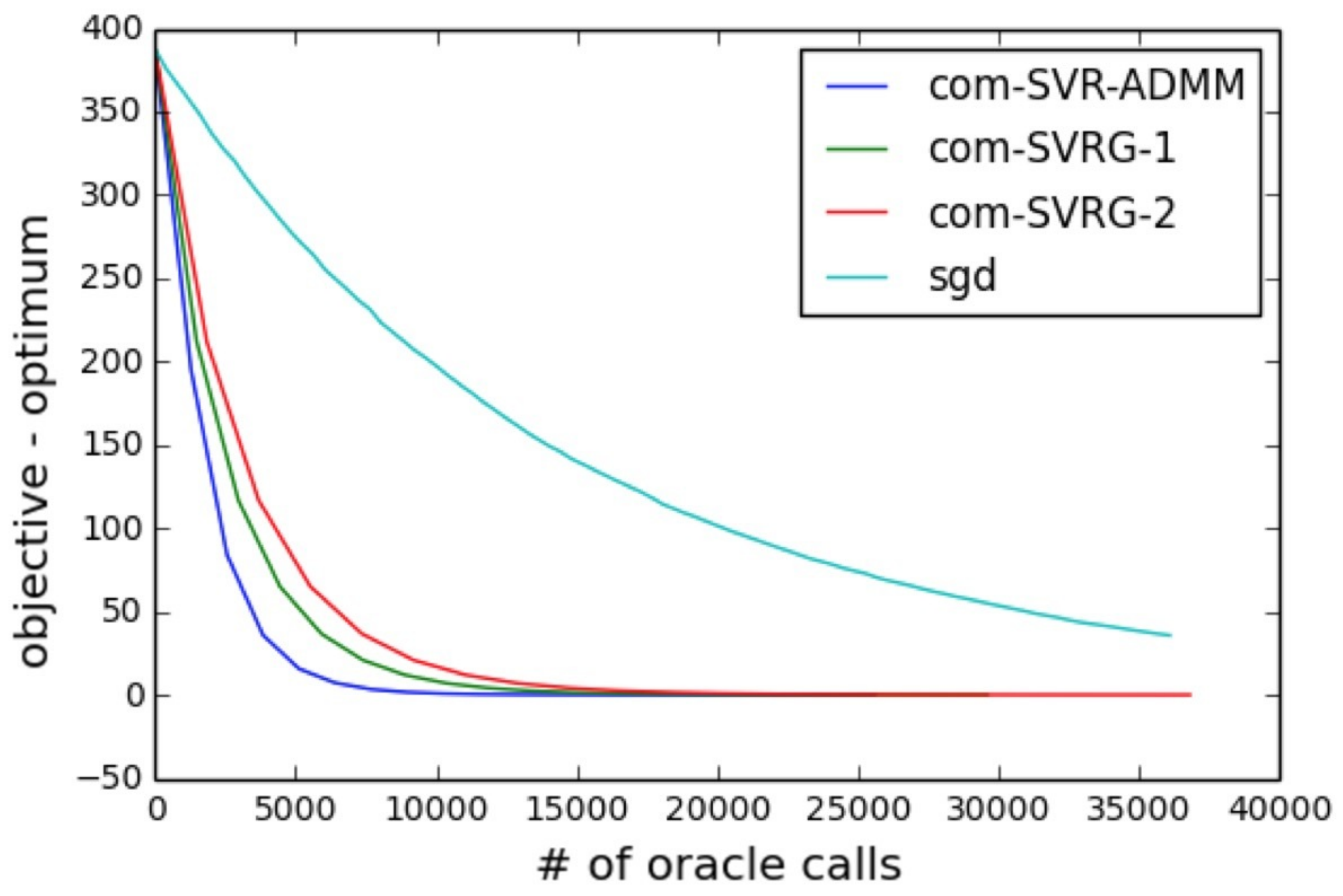}
  \label{fig:rl-2-1}
\endminipage\hfill
\minipage{0.46\textwidth}
  \includegraphics[width=\linewidth]{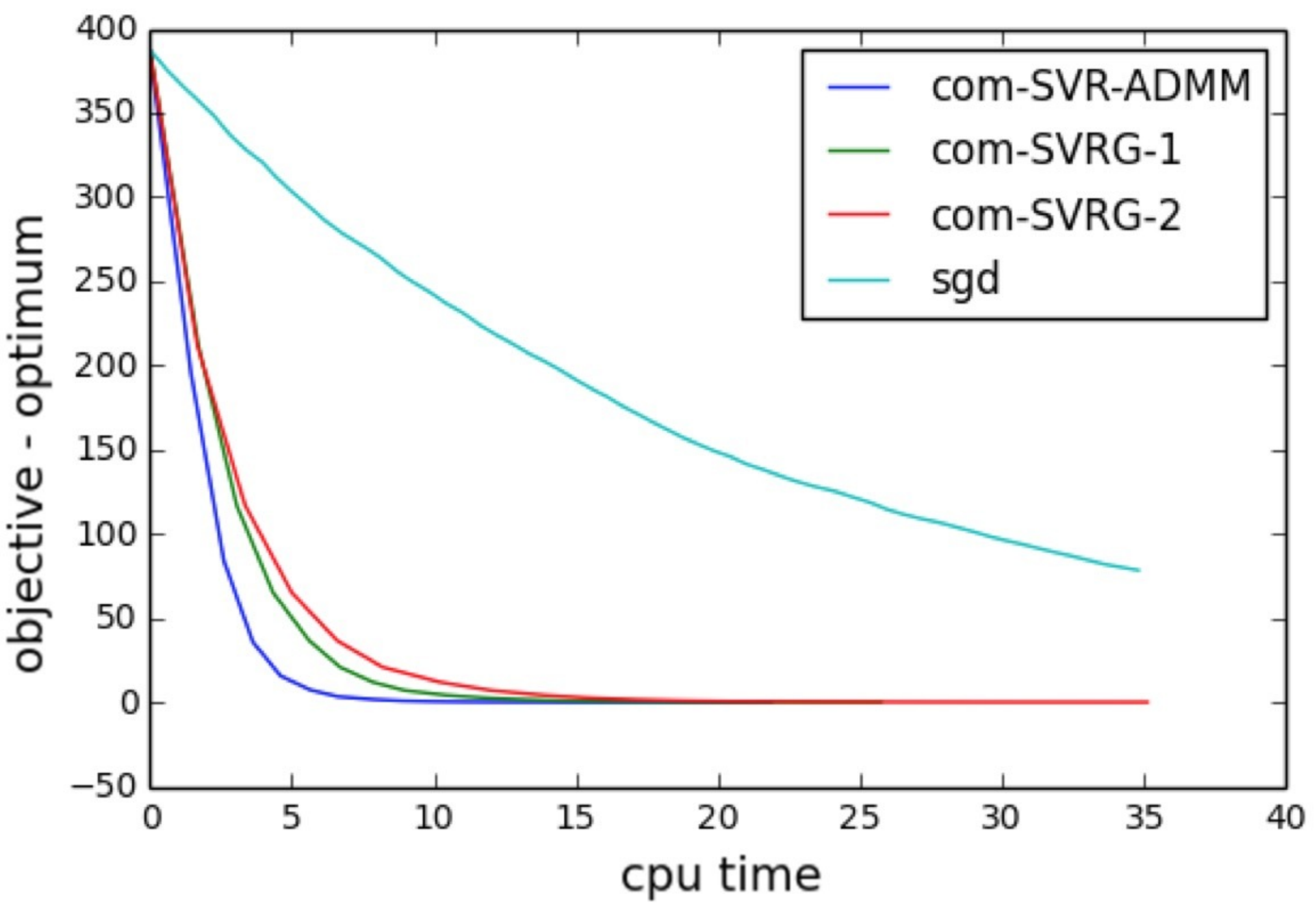}
  \label{fig:rl-2-2}
\endminipage
\caption{On-policy learning experiment with $S=200$, $d=200$. The other parameter values is the same as Figure ~\ref{rl-1}.}
\vskip -0.1in
\label{rl-2}
\end{figure}
\subsection{Reinforcement Learning}
Here we consider the problem (\ref{eq:rl-bell}), which can be used for
on-policy learning  \cite{wang2016accelerating}. In our experiment, we assume there are finite states and the number of states is  $S$. $\pi$ is the policy in consideration. $P_{s,s'}^\pi$ is the transition probability from state $s$ to $s'$ given policy $\pi$, $\gamma$ is a discount factor, $\phi_s \in \mathbb{R}^d$ is the feature of state $s$. Here we use a linear product $\langle\,\phi_s, w\rangle$ to approximate the value of state $s$. Our goal is to find the optimal  $w \in \mathbb{R}^d$.

We use the following specifications for oracles $g_{s'}(w)$ and $f_s(y)$:
\begin{eqnarray*}
 \hspace{-.3in} &&g_{s'}(w) = (\phi_1^Tw, r_{1,s'}+\gamma\phi_{s'}^Tw,  ...,\phi_S^Tw,  r_{S,s'}+\gamma\phi_{s'}^Tw)^{T}, \\
 \hspace{-.3in}  &&  f_{s}(y) = (y[2s-1] - y[2s])^2.
\end{eqnarray*}
Note here $g_{s'}(w)\in \mathbb{R}^{2S}$, and  $y[i]$ denote the $i$-th element of vector $y$.
All shared parameters in four algorithms have the same values. Note here that the calculation of $\mathbf{E}[g(w)]$ is no longer under uniform distribution. We use the given transition probability. In this experiment, the transition probability is randomly generated and then regularized. The reward is also randomly generated. In addition, we include a regularization term $R(w) = \frac{\mu}{2}||w||_2^2$ with $\mu > 0$. The results are shown in Figure~\ref{rl-1} and Figure~\ref{rl-2}. 
It can be seen that our proposed com-SVR-ADMM achieves faster convergence compared to the benchmark algorithms.

\section{Conclusion}
In this paper, we propose an ADMM-based algorithm, called com-SVR-ADMM, for stochastic composition optimization. We show that when the objective function is strongly convex and Lipschitz smooth, com-SVR-ADMM converges linearly.
In the case when the objective function is  convex (not necessarily strongly convex) and Lipschitz smooth, com-SVR-ADMM improves the theoretical convergence rate from $O(S^{-4/9})$ in \cite{wang2016accelerating} to $O(\frac{\log S}{S})$.
When the objective is only assumed to be convex, com-SVR-ADMM has a convergence rate of $O(\frac{1}{\sqrt{S}})$.
Experimental results show that com-SVR-ADMM outperforms existing algorithms.

\section*{Acknowledgements}
This work was supported in part by the National Natural Science Foundation of China Grants 61672316, 61303195, the Tsinghua Initiative Research Grant, and the China youth 1000-talent grant.


\bibliographystyle{abbrv}

\clearpage

\section{Appendix}

Recall that the stochastic composition problem we want to solve has the following form:\\
\begin{eqnarray}\label{eq:pro-form-1}
&\min_{x,\omega}& F(x) + R(\omega) \\
&\text{s.t.}&  Ax+B\omega = 0.\label{eq:pro-form-2}
\end{eqnarray}
where $F(x)\triangleq\frac{1}{n} \sum\limits_{i=1}^{n} f_{i}(\frac{1}{m}\sum\limits_{j=1}^{m}g_j(x))$. For clarity, we denote $F(x) = \frac{1}{n}\sum\limits_{i=1}^nF_i(x)$, $F_i(x) = f_i(g(x))$, $g(x) = \mathbf{E}g(x) = \frac{1}{m}\sum\limits_{j=1}^m g_j(x)$.
Therefore, $\nabla F_i(x) = (\partial g(x))^T \nabla f_i(g(x))$ and the augmented Lagrangian equation for (\ref{eq:pro-form-1}, \ref{eq:pro-form-2}) is
\begin{eqnarray}
  L_\rho(x,\omega,\lambda)  = F(x) + R(\omega) + \langle\, \lambda, Ax+B\omega\rangle\ + \frac{\rho}{2}||Ax + B\omega||_2^2, \quad \rho > 0.
\end{eqnarray}
Denote $(x^*, \omega^*, \lambda^*)$ as the optimal solution of (\ref{eq:pro-form-1}, \ref{eq:pro-form-2}), then it can be verified that the KKT conditions are
\begin{equation}
  \nabla F(x^*) = -A^T\lambda^*, \quad \tilde{\nabla}R(\omega^*) = -B^T\lambda^*, \quad Ax^*+B\omega^* = 0.
\end{equation}
Moreover, if matrix $A$ has full row rank, $\lambda^* = -(A^T)^{\dag}\nabla F(x^*)$ \cite{zheng2016fast}.
\\
\begin{proposition} Under Assumption \ref{assump:7}, we obtain \ $||\nabla F_i(x) - \nabla F_i(y)||
\leq L_F||x-y||$, $\forall x,y$.
\label{pro:F-lip}
\end{proposition}
\noindent
\emph{Proof.}
\begin{equation}
  \begin{aligned}
    ||\nabla F_i(x) - \nabla F_i(y)|| &= ||(\partial g(x))^T\nabla f_i(g(x)) - (\partial g(y))^T\nabla f_i(g(y))||\\
    &=||\frac{1}{m}\sum\limits_{j=1}^{m} ((\partial g_j(x))^T\nabla f_i(g(x)) - (\partial g_j(y))^T\nabla f_i(g(y)))||\\
    &\leq \frac{1}{m}\sum\limits_{j=1}^{m}||(\partial g_j(x))^T\nabla f_i(g(x)) - (\partial g_j(y))^T\nabla f_i(g(y))||\\
    &\leq L_F||x-y||
  \end{aligned}
\end{equation}
According to the definition of $F(x)$, it can be easily verified that $||\nabla F(x) - \nabla F(y)|| \leq L_F||x-y||$.
\QEDB
\\
\begin{proposition}
  \label{proposition:2}
Denote $\phi_{n_k} =[g_{n_k}(\tilde{x}) -  g_{n_k}(x^{k})]  - [g(\tilde{x}) - g(x^k)] $, then we have
\begin{equation}
  \mathbf{E}||\frac{1}{N}\sum_{n_k \in N_k} \phi_{n_k}||_2^2 = \frac{1}{N}\mathbf{E}||\phi_n||_2^2, \qquad \forall n \in {1,...,m}
\end{equation}
\end{proposition}
\noindent
\emph{Proof.}
\begin{equation}
  \begin{aligned}
  \mathbf{E}||\frac{1}{N}\sum_{n_k \in N_k} \phi_{n_k}||_2^2 &= \frac{1}{N^2}\mathbf{E}||\sum_{n_k \in N_k} \phi_{n_k}||_2^2\\
  &= \frac{1}{N^2}\mathbf{E}\sum_{n_k, n_k^{\prime} \in N_k} \phi_{n_k}^T\phi_{n_k^{\prime}}\\
  &=\frac{1}{N^2}\mathbf{E}\sum_{n_k\neq n_k^{\prime} \in N_k} \phi_{n_k}^T\phi_{n_k^{\prime}} + \frac{1}{N^2}\mathbf{E}\sum_{n_k \in N_k}||\phi_{n_k}||_2^2 \\
  &=\frac{1}{N}\mathbf{E}||\phi_n||_2^2
\end{aligned}
\end{equation}
where  we use $\mathbf{E}\sum_{n_k\neq n_k^{\prime} \in N_k} \phi_{n_k}^T\phi_{n_k^{\prime}} = 0$ in the last enquality\footnote{Because each element of $N_k$ is uniformly and indenpendently sampling from $\{1,...,m\}$ with replacement.} .
\QEDB
\\
\\
\\
\\
\noindent
\subsection{Proof of Theorem 1}
In this section, we prove the theoretical result shown in Theorem \ref{thm:1}. Denote $\mathbb{I}_k = \{i_k,j_k,N_k\}$.

\begin{lemma}  \label{lemma:1}$-B^{T}\lambda^{k} - \rho B^{T}(Ax^k + B\omega^{k+1}) \in \partial R(\omega^{k+1})$, where $\partial R(\omega)$ is the subdifferential of $R$ at point $\omega$.
\end{lemma}
\noindent
\emph{Proof.} The optimality condition of $w^{k+1}$'s update is:
\begin{equation}
  \label{eq:opt-omega}
0 \in \partial R(w^{k+1}) + B^{T}\lambda^{k} + \rho B^{T}(Ax^{k} + B\omega^{k+1})\\
\end{equation}
rearranging the terms  we obtain lemma \ref{lemma:1}.
\QEDB
\\
\\
\noindent
Now we transform the update of $x^{k+1}$ into the gradient descent form. The optimality condition of $x^{k+1}$'s update is:
\begin{equation}
  0 = \nabla\hat{F}_{i_k}(x^k) + A^{T}\lambda^{k} + \rho A^{T}(Ax^{k+1} + B\omega^{k+1}) + \frac{1}{\eta}(x^{k+1} - x^k)
\end{equation}
using $\lambda^{k+1} = \lambda^k + \rho(Ax^{k+1} + B\omega^{k+1})$ and rearranging terms, we have:
\begin{equation}
x^{k+1} = x^{k} - \eta(\nabla\hat{F}_{i_k}(x^k) + A^{T}\lambda^{k+1})
\end{equation}
Denoting $\mu_{i_k}^{k} = \nabla\hat{F}_{i_k}(x^k) + A^{T}\lambda^{k+1}$, we have:
\begin{equation}
  \label{eq:gra}
  x^{k+1} = x^{k} - \eta \mu_{i_k}^{k}
\end{equation}
\noindent
\begin{lemma}
  \label{lemma:2} Under Assumptions \ref{assump:1} and \ref{assump:7}, \
 if \ $0 \leq \eta \leq \frac{1}{L_F}$, we have \ $-2\eta \langle\,\mu_{i_k}^{k}, x^k - x^* \rangle + \eta^2||\mu_{i_k}^k||_2^2 \leq -2\eta(F(x^{k+1}) - F(x^*)) - 2\eta\langle\,\nabla\hat{F}_{i_k}(x^k)-\nabla F(x^k), x^{k+1} - x^* \rangle + 2\eta \langle\, A^{T}\lambda^{k+1}, x^* - x^{k+1}\rangle$
\end{lemma}
\noindent
\emph{Proof.} \ Because of the convexity of $F$ we have:
\begin{equation}
  \begin{aligned}
    \label{eq:F_first}
F(x^*) &\geq F(x^k) + \langle\, \nabla F(x^k), x^* - x^k\rangle\\
&\geq F(x^{k+1}) - \langle\, \nabla F(x^k), x^{k+1} - x^k\rangle - \frac{L_F}{2}||x^{k+1} - x^{k}||_2^2 + \langle\, \nabla F(x^k), x^* - x^k\rangle\\
& = F(x^{k+1}) - \langle\, \nabla F(x^k), x^{k+1} - x^k\rangle - \frac{L_F}{2}||\eta \mu_{i_k}^k||_2^2 +  \langle\, \nabla F(x^k), x^* - x^k\rangle\\
& = F(x^{k+1}) + \langle\, \nabla F(x^k), x^{*} - x^{k+1}\rangle - \frac{L_F}{2}\eta^2||\mu_{i_k}^k||_2^2
  \end{aligned}
\end{equation}
where in the second inequality, we use the Lipschitz smoothness of $F$ and use (\ref{eq:gra}) in the first equality.
\begin{equation}
  \begin{aligned}
    \label{eq:F_sec}
  \because  \ & \langle\, \nabla F(x^k), x^{*} - x^{k+1}\rangle + \langle\, A^{T}\lambda^{k+1}, x^{*} - x^{k+1}\rangle\\
  &=\langle\, \nabla F(x^k), x^{*} - x^{k+1}\rangle + \langle\, \mu_{i_k}^k - \nabla\hat{F}_{i_k}(x^k), x^{*} - x^{k+1}\rangle \\
  &=\langle\, \nabla F(x^k) - \nabla\hat{F}_{i_k}(x^k) , x^{*} - x^{k+1}\rangle + \langle\, \mu_{i_k}^k, x^{*} - x^{k+1}\rangle\\
  &=\langle\, \nabla F(x^k) - \nabla\hat{F}_{i_k}(x^k) , x^{*} - x^{k+1}\rangle  + \langle\, \mu_{i_k}^k, x^{*} -x^k + x^k- x^{k+1}\rangle\\
  &\overset{(\ref{eq:gra})}{=} \langle\, \nabla F(x^k) - \nabla\hat{F}_{i_k}(x^k) , x^{*} - x^{k+1}\rangle + \langle\, \mu_{i_k}^k, x^{*} -x^k\rangle + \langle\, \mu_{i_k}^k, \eta \mu_{i_k}^k\rangle
\end{aligned}
\end{equation}
Adding the both sides of (\ref{eq:F_first}) by $\langle\, A^{T}\lambda^{k+1}, x^{*} - x^{k+1}\rangle$ and using (\ref{eq:F_sec}), we have:
\begin{equation}
  \begin{aligned}
    \label{eq:eta}
  &F(x^*) +  \langle\, A^{T}\lambda^{k+1}, x^{*} - x^{k+1}\rangle \\
  & \geq F(x^{k+1}) + \langle\, \nabla F(x^k) - \nabla\hat{F}_{i_k}(x^k) , x^{*} - x^{k+1}\rangle + \langle\, \mu_{i_k}^k, x^{*} -x^k\rangle + \eta||\mu_{i_k}^k||_2^2 - \frac{L_F}{2}\eta^2||\mu_{i_k}^k||_2^2
\end{aligned}
\end{equation}
when $0\leq\eta\leq \frac{1}{L_F}$,\ $\eta - \frac{L_F}{2}\eta^2 \geq \frac{1}{2}\eta$. Therefore we have:
\begin{equation}
  \begin{aligned}
    \label{eq:eta_bound}
  &F(x^*) +  \langle\, A^{T}\lambda^{k+1}, x^{*} - x^{k+1}\rangle\\
  & \geq F(x^{k+1}) + \langle\, \nabla F(x^k) - \nabla\hat{F}_{i_k}(x^k) , x^{*} - x^{k+1}\rangle + \langle\, \mu_{i_k}^k, x^{*} -x^k\rangle + \frac{\eta}{2}||\mu_{i_k}^k||_2^2
  \end{aligned}
\end{equation}
Multiplying $2\eta$ of both sides and rearranging the terms we obtain Lemma \ref{lemma:2}.
\QEDB
\\
\begin{lemma} Under Assumptions \ref{assump:1}, \ref{assump:7}, \ref{assump:4} and \ref{assump:5}, \label{lemma:3} denote $\sigma(N) = \sqrt{\frac{1}{N}}$,
  \begin{equation}
    \begin{aligned}
      \label{eq:x}
   2\eta &\mathbf{E}[F(x^{k+1}) - F(x^*) - \langle\, \nabla F(x^*), x^{k+1}-x^* \rangle + \langle\, \lambda^{k+1} - \lambda^{*}, Ax^{k+1}-Ax^* \rangle] \\
    \leq &\mathbf{E}||x^{k} - x^*||_2^2 - \mathbf{E}||x^{k+1} - x^*||_2^2 + (\frac{16\eta^2C_G^4L_f^2}{ N} +24\eta^2L_F^2+4\eta DC_GL_fL_G\sigma(N)) \mathbf{E}||x^k - x^*||_2^2 \\
  &+ (\frac{16\eta^2C_G^4L_f^2}{ N} +24\eta^2L_F^2+4\eta DC_GL_fL_G\sigma(N)) ||\tilde{x} - x^*||_2^2
  \end{aligned}
  \end{equation}
\end{lemma}
\noindent
\emph{Proof.} \
Both sides of (\ref{eq:gra}) minus $x^*$ to yield:
\begin{equation}
  \begin{aligned}
  ||x^{k+1} - x^*||_2^2 &= ||x^{k} - x^* - \eta\mu_{i_k}^{k}||_2^2\\
                        &= ||x^{k} - x^*||_2^2 - 2\eta \langle\, \mu_{i_k}^{k}, x^{k} - x^*\rangle  + \eta^{2} ||\mu_{i_k}^{k}||_2^2
\end{aligned}
\end{equation}
using Lemma \ref{lemma:2} we obtain:
\begin{equation}
  \begin{aligned}
    ||x^{k+1} - x^*||_2^2 \leq & ||x^{k} - x^*||_2^2 - 2\eta(F(x^{k+1}) - F(x^*)) - 2\eta\langle\,\nabla\hat{F}_{i_k}(x^k)-\nabla F(x^k), x^{k+1} - x^* \rangle \\
    & + 2\eta \langle\, A^{T}\lambda^{k+1}, x^* - x^{k+1}\rangle
  \end{aligned}
\end{equation}
Rearranging terms we have:
\begin{equation}
  \begin{aligned}
2\eta(F(x^{k+1}) - F(x^*)) - 2\eta \langle\, A^{T}\lambda^{k+1}, x^* - x^{k+1}\rangle \leq & ||x^{k} - x^*||_2^2 - ||x^{k+1} - x^*||_2^2 \\
&  - 2\eta\langle\,\nabla\hat{F}_{i_k}(x^k)-\nabla F(x^k), x^{k+1} - x^* \rangle
\end{aligned}
\end{equation}
Taking expectation w.r.t. $\mathbb{I}_k$ in the current step $s$, we have:
\begin{equation}
  \begin{aligned}
    \label{eq: T_1}
  2\eta \mathbf{E}[F(x^{k+1}) - F(x^*)] - 2\eta \mathbf{E}[ \langle\, A^{T}\lambda^{k+1}, x^* - x^{k+1}\rangle] &\leq \mathbf{E} ||x^{k} - x^*||_2^2 - \mathbf{E}||x^{k+1} - x^*||_2^2\\
  & \underbrace{ - 2\eta \mathbf{E}[\langle\,\nabla\hat{F}_{i_k}(x^k)-\nabla F(x^k), x^{k+1} - x^* \rangle] }_{T_1}
  \end{aligned}
\end{equation}
Now we bound $T_1$. Using $(19,20)$ in the proof of Theorem 1 in \cite{zheng2016fast} we have:
\begin{equation}
  \label{eq:ref-1}
  - 2\eta \langle\,\nabla\hat{F}_{i_k}(x^k)-\nabla F(x^k), x^{k+1} - x^* \rangle \leq 2\eta^2||\nabla\hat{F}_{i_k}(x^k) - \nabla F(x^k) ||_2^2 - 2\eta \langle\, \nabla\hat{F}_{i_k}(x^k) -\nabla F(x^k) , \bar{x} - x^{*}\rangle
\end{equation}
where
\begin{equation}
  \begin{aligned}
  \bar{x} &= prox_{\eta \phi_{k+1} } (x^k - \eta\nabla F(x^k))\\
   \phi_{k+1}(x) &= \langle\, \lambda^k, Ax\rangle + \frac{\rho}{2}||Ax + B\omega^{k+1}||_2^2 + \frac{1}{2\eta}||x-x^k||_{G_k - I}^2
\end{aligned}
\end{equation}
Taking expectation of $\mathbb{I}_k$ on (\ref{eq:ref-1}), we have:
\begin{equation}
\begin{aligned}
  T_1 & \leq 2\eta^2\mathbf{E}||\nabla\hat{F}_{i_k}(x^k) - \nabla F(x^k) ||_2^2 - 2\eta \mathbf{E}\langle\, \nabla\hat{F}_{i_k}(x^k) -\nabla F(x^k) , \bar{x} - x^{*}\rangle\\
&  \leq  2\eta^2\mathbf{E}||\nabla\hat{F}_{i_k}(x^k) - \nabla F(x^k) ||_2^2 -2\eta\mathbf{E} \langle\,  (\partial g_{j_k}(x^k))^{T}\nabla f_{i_k}(\hat{g}(x^k))
-(\partial g_{j_k}(\tilde{x}) )^{T}\nabla f_{i_k}(g(\tilde{x})) +\nabla F(\tilde{x}) -\nabla F(x^k),\bar{x} - x^{*}\rangle\\
& =   2\eta^2\mathbf{E}||\nabla\hat{F}_{i_k}(x^k) - \nabla F(x^k) ||_2^2 -2\eta\mathbf{E} \langle\,  (\partial g_{j_k}(x^k))^{T}\nabla f_{i_k}(\hat{g}(x^k)) -\nabla F(x^k),\bar{x} - x^{*}\rangle\\
& =   2\eta^2\mathbf{E}||\nabla\hat{F}_{i_k}(x^k) - \nabla F(x^k) ||_2^2 -2\eta\mathbf{E} \langle\,  (\partial g_{j_k}(x^k))^{T}\nabla f_{i_k}(\hat{g}(x^k)) -(\partial g_{j_k}(x^k))^{T}\nabla f_{i_k}(g(x^k)),\bar{x} - x^{*}\rangle\\
& \leq 2\eta^2\underbrace{\mathbf{E}||\nabla\hat{F}_{i_k}(x^k) - \nabla F(x^k) ||_2^2}_{T_2} + 2\eta D\underbrace{\mathbf{E}||(\partial g_{j_k}(x^k))^{T}\nabla f_{i_k}(\hat{g}(x^k)) -(\partial g_{j_k}(x^k))^{T}\nabla f_{i_k}(g(x^k))||_2}_{T_3}
\end{aligned}
\end{equation}
where we use Assumption \ref{assump:1}$(iii)$ in the first and second equality, therefore
\begin{equation}
  T_1 \leq 2\eta^2 T_2 + 2\eta DT_3
\end{equation}
Now we bound $T_2$.
\begin{equation}
  \begin{aligned}
    T_2 &= \mathbf{E}||\nabla\hat{F}_{i_k}(x^k) - \nabla F(x^k) ||_2^2\\
        &= \mathbf{E}|| (\partial g_{j_k}(x^k))^{T}\nabla f_{i_k}(\hat{g}(x^k))-(\partial g_{j_k}(\tilde{x}) )^{T}\nabla f_{i_k}(g(\tilde{x})) +\nabla F(\tilde{x}) -\nabla F(x^k)  ||_2^2\\
        &\leq 2\mathbf{E}|| (\partial g_{j_k}(x^k))^{T}\nabla f_{i_k}(\hat{g}(x^k))-(\partial g_{j_k}(\tilde{x}) )^{T}\nabla f_{i_k}(g(\tilde{x}))||_2^2 + 2\mathbf{E}||\nabla F(\tilde{x}) -\nabla F(x^k)||_2^2\\
        &= 2\mathbf{E}|| (\partial g_{j_k}(x^k))^{T}\nabla f_{i_k}(\hat{g}(x^k))-(\partial g_{j_k}(x^k))^{T}\nabla f_{i_k}(g(x^k)) + (\partial g_{j_k}(x^k))^{T}\nabla f_{i_k}(g(x^k)) - (\partial g_{j_k}(\tilde{x}) )^{T}\nabla f_{i_k}(g(\tilde{x}))||_2^2 \\
        &\qquad  + 2\mathbf{E}||\nabla F(\tilde{x}) -\nabla F(x^k)||_2^2\\
        &\leq 4\mathbf{E}||(\partial g_{j_k}(x^k))^{T}\nabla f_{i_k}(\hat{g}(x^k))-(\partial g_{j_k}(x^k))^{T}\nabla f_{i_k}(g(x^k))||_2^2 + 4\mathbf{E}||(\partial g_{j_k}(x^k))^{T}\nabla f_{i_k}(g(x^k)) - (\partial g_{j_k}(\tilde{x}) )^{T}\nabla f_{i_k}(g(\tilde{x}))||_2^2 \\
        & \qquad + 2L_F^2\mathbf{E}||\tilde{x}-x^k||_2^2\\
        & \leq 4\mathbf{E}||(\partial g_{j_k}(x^k))^{T}\nabla f_{i_k}(\hat{g}(x^k))-(\partial g_{j_k}(x^k))^{T}\nabla f_{i_k}(g(x^k))||_2^2 + 4L_F^2\mathbf{E}||\tilde{x} - x^k||_2^2 + 2L_F^2\mathbf{E}||\tilde{x} - x^k||_2^2\\
        &\leq 4 \underbrace{\mathbf{E}||(\partial g_{j_k}(x^k))^{T}\nabla f_{i_k}(\hat{g}(x^k))-(\partial g_{j_k}(x^k))^{T}\nabla f_{i_k}(g(x^k))||_2^2}_{T_4} + 12L_F^2||\tilde{x} - x^*||_2^2 + 12L_F^2\mathbf{E}||x^k - x^*||_2^2
  \end{aligned}
\end{equation}
where we use Proposition \ref{pro:F-lip} in the second inequality and use Assumption \ref{assump:7} in the third inequality. We then bound $T_4$.
\begin{equation}
  \begin{aligned}
    T_4 = &\mathbf{E}||(\partial g_{j_k}(x^k))^{T}\nabla f_{i_k}(\hat{g}(x^k))- (\partial g_{j_k}(x^k))^T\nabla f_{i_k}(g(x^k))||_2^2\\
        = &\mathbf{E}||(\partial g_{j_k}(x^k))^{T}[\nabla f_{i_k}(\hat{g}(x^k))-\nabla f_{i_k}(g(x^k))]||_2^2\\
        \leq & \mathbf{E}(||\partial g_{j_k}(x^k)||_2^2 ||\nabla f_{i_k}(\hat{g}(x^k))-\nabla f_{i_k}(g(x^k))||_2^2)\\
        \leq& C_G^{2}\mathbf{E} ||\nabla f_{i_k}(\hat{g}(x^k))-\nabla f_{i_k}(g(x^k))||_2^2 \quad \quad (Assumption \ \ref{assump:5})\\
        \leq & C_G^{2} L_f^{2}\mathbf{E}||\hat{g}(x^{k}) - g(x^{k})||_2^2 \quad \quad (Assumption \ \ref{assump:4})\\
        = & C_G^{2} L_f^{2} \mathbf{E}||g(\tilde{x}) - \frac{1}{N} \sum\limits_{1\leq j \leq N} \left( g_{N_k[j]}(\tilde{x}) -  g_{N_k[j]}(x^{k})\right) - g(x^k)||_2^2\\
        = & C_G^{2} L_f^{2} \mathbf{E}||\frac{1}{N} \sum\limits_{1\leq j \leq N} \{ [g_{N_k[j]}(\tilde{x}) -  g_{N_k[j]}(x^{k})]  - [g(\tilde{x}) - g(x^k)] \} ||_2^2\\
        =& \frac{ C_G^{2} L_f^{2} } {N^2} \mathbf{E}||\sum\limits_{1\leq j \leq N} \{ [g_{N_k[j]}(\tilde{x}) -  g_{N_k[j]}(x^{k})]  - [g(\tilde{x}) - g(x^k)] \}||_2^2\\
        =& \frac{ C_G^{2} L_f^{2} } {N^2}\sum\limits_{1\leq j \leq N} \mathbf{E} ||[g_{N_k[j]}(\tilde{x}) -  g_{N_k[j]}(x^{k})]  - [g(\tilde{x}) - g(x^k)]||_2^2\\
        \leq& \frac{C_G^{2} L_f^{2}}{N^2}\sum\limits_{1\leq j \leq N} \mathbf{E} ||g_{N_k[j]}(\tilde{x}) -  g_{N_k[j]}(x^{k}) ||_2^2 \quad \quad (\mathbf{E}||x - \mathbf{E}x||_2^2 \leq \mathbf{E}||x||_2^2)\\
        =& \frac{C_G^{2} L_f^{2}}{N^2}\sum\limits_{1\leq j \leq N} \mathbf{E}||g_{N_k[j]}(\tilde{x}) - g_{N_k[j]}(x^*) + g_{N_k[j]}(x^*) - g_{N_k[j]}(x^{k}) ||_2^2\\
        \leq & \frac{2C_G^{2} L_f^{2}}{N^2}\sum\limits_{1\leq j \leq N} \mathbf{E}||g_{N_k[j]}(\tilde{x}) - g_{N_k[j]}(x^*)||_2^2 + \frac{2C_G^{2} L_f^{2}}{N^2}\sum\limits_{1\leq j \leq N} \mathbf{E}||g_{N_k[j]}(x^*) - g_{N_k[j]}(x^{k}) ||_2^2\\
        \leq & \frac{2C_G^{4} L_f^{2}}{N^2}\sum\limits_{1\leq j \leq N} \mathbf{E}||x^k - x^*||_2^2 + \frac{2C_G^{4} L_f^{2}}{N^2}\sum\limits_{1\leq j \leq N} ||\tilde{x} - x^*||_2^2\\
        = & \frac{2C_G^{4} L_f^{2}}{N} \mathbf{E}||x^k - x^*||_2^2 + \frac{2C_G^{4} L_f^{2}}{N} ||\tilde{x} - x^*||_2^2
  \end{aligned}
\end{equation}
where the sixth equality is from equation $(27)$ in \cite{lian2016finite}.
\begin{equation}
  \begin{aligned}
    \therefore  \ \ T_2 \leq & 4 T_4 +  + 12L_F^2||\tilde{x} - x^*||^2 + 12L_F^2\mathbf{E}||x^k - x^*||^2\\
                        \leq & 4[\frac{2C_G^{4} L_f^{2}}{N}\mathbf{E} ||x^k - x^*||_2^2 + \frac{2C_G^{4} L_f^{2}}{N} ||\tilde{x} - x^*||_2^2] + 12L_F^2||\tilde{x} - x^*||^2 + 12L_F^2\mathbf{E}||x^k - x^*||^2\\
                        = & (\frac{8 C_G^{4} L_f^{2}}{ N} + 12L_F^2) \mathbf{E}||x^k - x^*||_2^2 + (\frac{8 C_G^{4} L_f^{2}}{ N} + 12L_F^2) ||\tilde{x} - x^*||_2^2
  \end{aligned}
\end{equation}
Now we bound $T_3$.
\begin{equation}
  \begin{aligned}
  T_3 &= \mathbf{E}||(\partial g_{j_k}(x^k))^{T}\nabla f_{i_k}(\hat{g}(x^k)) -(\partial g_{j_k}(x^k))^{T}\nabla f_{i_k}(g(x^k))||_2\\
      &\leq C_G\mathbf{E}||\nabla f_{i_k}(\hat{g}(x^k)) -\nabla f_{i_k}(g(x^k))||_2\\
      &\leq C_GL_f\mathbf{E}||\hat{g}(x^k) - g(x^k)||_2\\
      & = C_GL_f\mathbf{E}||g(\tilde{x}) - \frac{1}{N} \sum\limits_{1\leq j \leq N} \left( g_{\mathbb{N}_k[j]}(\tilde{x}) -  g_{\mathbb{N}_k[j]}(x^{k})\right) - g(x^k)||_2\\
      &= C_GL_f\mathbf{E}||\frac{1}{N} \sum\limits_{1\leq j \leq N} \{ [g_{N_k[j]}(\tilde{x}) -  g_{N_k[j]}(x^{k})]  - [g(\tilde{x}) - g(x^k)] \} ||_2\\
      &= C_GL_f\mathbf{E}||\frac{1}{N} \sum\limits_{n_k \in N_k} \{ [g_{n_k}(\tilde{x}) -  g_{n_k}(x^{k})]  - [g(\tilde{x}) - g(x^k)] \} ||_2\\
    \end{aligned}
\end{equation}
Recall the definition of $\phi_{n_k}$ in Proposition \ref{proposition:2}, we obtain
\begin{equation}
  \begin{aligned}
  T_3 &= C_GL_f\mathbf{E}||\frac{1}{N} \sum\limits_{n_k \in N_k} \phi_{n_k}||_2\\
  &\leq C_GL_f(\mathbf{E}||\frac{1}{N} \sum\limits_{n_k \in N_k} \phi_{n_k}||_2^2)^{1/2}\\
  &=C_GL_f(\frac{1}{N}\mathbf{E}||\phi_n||_2^2)^{1/2}
\end{aligned}
\end{equation}
where we use $(\mathbf{E}X)^2 \leq \mathbf{E}X^2$ in the first inequality and using Proposition \ref{proposition:2} in the second equality. Now lets bound $\mathbf{E}||\phi_n||_2^2$.
\begin{equation}
  \begin{aligned}
    \mathbf{E}||\phi_n||_2^2 &= \mathbf{E}||[g_{n}(\tilde{x}) -  g_{n}(x^{k})]  - [g(\tilde{x}) - g(x^k)]||_2^2\\
    &\leq \mathbf{E}||g_{n}(\tilde{x}) -  g_{n}(x^{k})||_2^2\\
    &\leq L_G^2\mathbf{E}||x^k - \tilde{x}||_2^4
  \end{aligned}
\end{equation}
where the last inequaliy is from Assumption \ref{assump:5}. Therefore
\begin{equation}
  \begin{aligned}
  T_3 &\leq C_GL_fL_G\sqrt{\frac{1}{N}}\mathbf{E}||x^k-\tilde{x}||^2\\
  &\leq 2C_GL_fL_G\sigma(N)\mathbf{E}||x^k-x^*||^2 + 2C_GL_fL_G\sigma(N)||\tilde{x}-x^*||^2
\end{aligned}
\end{equation}
\begin{equation}
  \begin{aligned}
  \therefore \ T_1&\leq 2\eta^2 T_2 + 2\eta DT_3\\
  & = (\frac{16\eta^2C_G^4L_f^2}{N} + 24\eta^2L_F^2 + 4\eta DC_GL_fL_G\sigma(N))\mathbf{E}||x^k - x^*||^2 \\
  & \qquad + (\frac{16\eta^2C_G^4L_f^2}{N} + 24\eta^2L_F^2 + 4\eta DC_GL_fL_G\sigma(N))||\tilde{x} - x^*||^2
\end{aligned}
\end{equation}
Putting back into (\ref{eq: T_1}) we obtain:
\begin{equation}
  \begin{aligned}
& \underbrace{2\eta \mathbf{E}[F(x^{k+1}) - F(x^*) + \langle\, A^{T}\lambda^{k+1}, x^{k+1} - x^*\rangle]}_{T_5} \leq \\
&  \mathbf{E}||x^{k} - x^*||_2^2 - \mathbf{E}||x^{k+1} - x^*||_2^2 + (\frac{16\eta^2C_G^4L_f^2}{N} + 24\eta^2L_F^2 + 4\eta DC_GL_fL_G\sigma(N))\mathbf{E}||x^k - x^*||^2 \\
& \qquad + (\frac{16\eta^2C_G^4L_f^2}{N} + 24\eta^2L_F^2 + 4\eta DC_GL_fL_G\sigma(N))||\tilde{x} - x^*||^2
\end{aligned}
\end{equation}
Now we turn to bound $T_5$.
\begin{equation}
  \begin{aligned}
    \label{eq:F_left}
    T_5 =& 2\eta \mathbf{E}[F(x^{k+1}) - F(x^*) + \langle\, A^{T}\lambda^{k+1}, x^{k+1} - x^*\rangle]\\
        =& 2\eta \mathbf{E}[F(x^{k+1}) - F(x^*) - \langle\, \nabla F(x^*), x^{k+1}-x^* \rangle + \langle\, \nabla F(x^*), x^{k+1}-x^* \rangle + \langle\, A^{T}\lambda^{k+1}, x^{k+1} - x^*\rangle] \\
        =& 2\eta \mathbf{E}[F(x^{k+1}) - F(x^*) - \langle\, \nabla F(x^*), x^{k+1}-x^* \rangle + \langle\, -A^{T}\lambda^{*}, x^{k+1}-x^* \rangle + \langle\, A^{T}\lambda^{k+1}, x^{k+1} - x^*\rangle] \\
        =& 2\eta \mathbf{E}[F(x^{k+1}) - F(x^*) - \langle\, \nabla F(x^*), x^{k+1}-x^* \rangle + \langle\, \lambda^{k+1} - \lambda^{*}, Ax^{k+1}-Ax^* \rangle]
  \end{aligned}
\end{equation}
where we use $\nabla F(x^*) = -A^T\lambda^*$ in the third equality. Therefore we have:
\begin{equation}
  \begin{aligned}
  & 2\eta \mathbf{E}[F(x^{k+1}) - F(x^*) - \langle\, \nabla F(x^*), x^{k+1}-x^* \rangle + \langle\, \lambda^{k+1} - \lambda^{*}, Ax^{k+1}-Ax^* \rangle]\\
  & \leq    \mathbf{E}||x^{k} - x^*||_2^2 - \mathbf{E}||x^{k+1} - x^*||_2^2 + (\frac{16\eta^2C_G^4L_f^2}{N} + 24\eta^2L_F^2 + 4\eta DC_GL_fL_G\sigma(N))\mathbf{E}||x^k - x^*||^2 \\
  & \qquad + (\frac{16\eta^2C_G^4L_f^2}{N} + 24\eta^2L_F^2 + 4\eta DC_GL_fL_G\sigma(N))||\tilde{x} - x^*||^2
  \end{aligned}
  \end{equation}
\QEDB
\noindent
\begin{lemma}
  \label{lemma:4}
\begin{equation}
  \begin{aligned}
    \label{eq:omega}
    & \mathbf{E}(R(\omega^{k+1}) - R(\omega^*) - \langle\, \tilde{\nabla}R(\omega^*), \omega^{k+1} - \omega^*\rangle + \langle\, \lambda^{k+1} - \lambda^*, B\omega^{k+1}-B\omega^*\rangle )\\
    & \leq \frac{\rho}{2}\mathbf{E} (||Ax^k + B\omega^*||_2^2 - ||Ax^{k+1}+B\omega^*||_2^2 + ||Ax^{k+1}+B\omega^{k+1}||_2^2)
  \end{aligned}
\end{equation}
\end{lemma}
\noindent
\emph{Proof.}
\begin{equation}
  \begin{aligned}
    \because R(\omega^{*}) \geq R(\omega^{k+1}) & + \langle\, \tilde{\nabla}R(\omega^{k+1}), \omega^* - \omega^{k+1} \rangle \ \  \quad (convexity)\\
    \therefore R(\omega^{k+1}) - R(\omega^*) & \leq \langle\, \tilde{\nabla}R(\omega^{k+1}), \omega^{k+1} - \omega^{*} \rangle\\
    & = \langle\, -B^{T}\lambda^{k} - \rho B^{T}(Ax^k + B\omega^{k+1}), \omega^{k+1} - \omega^*\rangle \quad (lemma \ \ref{lemma:1})\\
    &= \langle\, \lambda^{k+1} + \rho(Ax^k - Ax^{k+1}), B\omega^* - B\omega^{k+1}\rangle\\
    & = \langle\, \lambda^{k+1}, B\omega^* - B\omega^{k+1}\rangle + \rho\langle\,Ax^k - Ax^{k+1}, B\omega^* - B\omega^{k+1}\rangle\\
\end{aligned}
\end{equation}
Rearranging terms we have:
\begin{equation}
\begin{aligned}
  \label{eq:R_1}
&  \underbrace{R(\omega^{k+1}) - R(\omega^*) + \langle\, \lambda^{k+1}, B\omega^{k+1} - B\omega^{*}\rangle}_{T_6} \\
  & \leq \rho \langle\,(Ax^k - Ax^{k+1}), B\omega^* - B\omega^{k+1}\rangle\\
 &= \frac{\rho}{2}[||Ax^k + B\omega^*||_2^2 - ||Ax^k + B\omega^{k+1}||_2^2 + ||Ax^{k+1}+B\omega^{k+1}||_2^2 - ||Ax^{k+1}+B\omega^{*}||_2^2]\\
 &\leq \frac{\rho}{2}[||Ax^k + B\omega^*||_2^2 - ||Ax^{k+1}+B\omega^{*}||_2^2 + ||Ax^{k+1}+B\omega^{k+1}||_2^2]
\end{aligned}
\end{equation}
Then we rewritten $T_6$ as:
\begin{equation}
  \begin{aligned}
    \label{eq:R_2}
    T_6 &= R(\omega^{k+1}) - R(\omega^*) + \langle\, \lambda^{k+1}, B\omega^{k+1} - B\omega^{*}\rangle\\
        &= R(\omega^{k+1}) - R(\omega^*) - \langle\, \tilde{\nabla}R(\omega^*), \omega^{k+1} - \omega^{*}\rangle + \langle\, \tilde{\nabla}R(\omega^*), \omega^{k+1} - \omega^{*}\rangle +  \langle\, \lambda^{k+1}, B\omega^{k+1} - B\omega^{*}\rangle\\
        &= R(\omega^{k+1}) - R(\omega^*) - \langle\, \tilde{\nabla}R(\omega^*), \omega^{k+1} - \omega^{*}\rangle + \langle\, -B^{T}\lambda^*, \omega^{k+1} - \omega^{*}\rangle +  \langle\, \lambda^{k+1}, B\omega^{k+1} - B\omega^{*}\rangle\\
        &=R(\omega^{k+1}) - R(\omega^*) - \langle\, \tilde{\nabla}R(\omega^*), \omega^{k+1} - \omega^{*}\rangle  + \langle\, \lambda^{k+1} - \lambda^*, B\omega^{k+1} - B\omega^{*}\rangle
  \end{aligned}
\end{equation}
where the third equality uses $\tilde{\nabla}R(\omega^*) = -B^T\lambda^*$. Putting (\ref{eq:R_2}) into (\ref{eq:R_1}) we get Lemma $\ref{lemma:4}$.
\QEDB
\\
\\
\noindent
\begin{lemma}
  \label{lemma:lambda}
\begin{equation}
  \begin{aligned}
    \label{eq:lambda}
    -\mathbf{E}\langle\, Ax^{k+1} + B\omega^{k+1}, \lambda^{k+1} - \lambda^*\rangle = \frac{1}{2\rho}\mathbf{E}(||\lambda^k - \lambda^*||_2^2 - ||\lambda^{k+1} - \lambda^*||_2^2 - ||\lambda^{k} - \lambda^{k+1}||_2^2)
    \end{aligned}
    \end{equation}
  \end{lemma}
  \noindent
\emph{Proof.} \ Using the update equation of $\lambda^{k+1}$ we have $Ax^{k+1} + B\omega^{k+1} = \frac{\lambda^{k+1} - \lambda^{k}}{\rho}$, then it is easy to verify (\ref{eq:lambda}).
\QEDB
\\
\\
\noindent
\emph{Proof \ of \ Theorem  \ref{thm:1}.} \
 Calculating $(\ref{eq:x}) + (\ref{eq:omega}) \times 2\eta + (\ref{eq:lambda})\times 2\eta$ and using $Ax^{k+1} + B\omega^{k+1} = \frac{\lambda^{k+1} - \lambda^{k}}{\rho}, \  Ax^* + B\omega^* = 0$, we have:
\begin{equation}
\begin{aligned}
  \label{eq:i16}
  &2\eta \mathbf{E} (F(x^{k+1}) - F(x^*) - \langle\, \nabla F(x^*), x^{k+1}-x^*\rangle + R(\omega^{k+1}) - R(\omega^*) - \langle\, \tilde{\nabla}R(x^*), \omega^{k+1}-\omega^*\rangle) \\
  &\leq \mathbf{E} (||x^k - x^*||_2^2 - ||x^{k+1} - x^*||_2^2) + (\frac{16\eta^2C_G^4L_f^2}{ N} +24\eta^2L_F^2+4\eta DC_GL_fL_G\sigma(N)) \mathbf{E}||x^k - x^*||_2^2 \\
&+ (\frac{16\eta^2C_G^4L_f^2}{ N} +24\eta^2L_F^2+4\eta DC_GL_fL_G\sigma(N)) ||\tilde{x} - x^*||_2^2 + \eta\rho \mathbf{E}(||Ax^{k} + B\omega^{*}||_2^2 - ||Ax^{k+1} + B\omega^{*}||_2^2)\\
  &+\frac{\eta}{\rho}\mathbf{E}(||\lambda^k - \lambda^*||_2^2 - ||\lambda^{k+1} - \lambda^*||_2^2)
\end{aligned}
\end{equation}
Because $F(x)$ is strongly convex with parameter $\mu_F$, namely,
\begin{equation}
  F(x) - F(x^*) \geq \langle\, \nabla F(x^*), x-x^*\rangle + \frac{\mu_F}{2}||x - x^*||_2^2,  \quad \forall x
\end{equation}
therefore
\begin{equation}
  ||x-x^{*}||_2^2 \leq \frac{2}{\mu_F}(F(x) - F(x^*) - \langle\, \nabla F(x^*), x-x^* \rangle), \quad \forall x
\end{equation}
According to the convexity of $R(\omega)$ we have for all $\omega$:
\begin{equation}
 R(\omega) - R(\omega^{*}) - \langle\, \tilde{\nabla}R(\omega^*), \omega - \omega^* \rangle \geq 0
 \end{equation}
 Recall the definition of $G(u)$, for all $x, \omega$, we further obtain:
\begin{equation}
  \label{eq:strongly}
  ||x-x^{*}||_2^2 \leq \frac{2}{\mu_F}(F(x) - F(x^*) - \langle\, \nabla F(x^*), x-x^* \rangle + R(\omega) - R(\omega^{*}) - \langle\, \tilde{\nabla}R(\omega^*), \omega - \omega^* \rangle) = \frac{2}{\mu_F}G(u)
\end{equation}
Therefore $(\ref{eq:i16})$ becomes:
\begin{equation}
  \begin{aligned}
2\eta \mathbf{E}(G(u^{k+1})) &\leq \mathbf{E}(||x^k - x^*||_2^2 - ||x^{k+1} - x^*||_2^2)\\
                            &+(\frac{32\eta^2C_G^4L_f^2}{\mu_F N} +\frac{48\eta^2L_F^2+8\eta DC_GL_fL_G\sigma(N)}{\mu_F})\mathbf{E}(G(u^k)) \\
                            &+ (\frac{32\eta^2C_G^4L_f^2}{\mu_F N} +\frac{48\eta^2L_F^2+8\eta DC_GL_fL_G\sigma(N)}{\mu_F}) G(\tilde{u})\\
                            &+\eta\rho \mathbf{E}(||Ax^{k} + B\omega^{*}||_2^2 - ||Ax^{k+1} + B\omega^{*}||_2^2)\\
                            &+\frac{\eta}{\rho}\mathbf{E}(||\lambda^k - \lambda^*||_2^2 - ||\lambda^{k+1} - \lambda^*||_2^2)
  \end{aligned}
\end{equation}
Summing from $k = 0,...,K-1$, we obtain:
\begin{equation}
\begin{aligned}
  \label{eq:i1}
2\eta \mathbf{E}\sum\limits_{k=0}^{K-1}(G(u^{k+1})) &\leq ||x^0 - x^*||_2^2\\
                                                    & + (\frac{32\eta^2C_G^4L_f^2}{\mu_F N} +\frac{48\eta^2L_F^2+8\eta DC_GL_fL_G\sigma(N)}{\mu_F})\mathbf{E}(\sum\limits_{k=0}^{K-1}G(u^k))\\
                                                    & + K(\frac{32\eta^2C_G^4L_f^2}{\mu_F N} +\frac{48\eta^2L_F^2+8\eta DC_GL_fL_G\sigma(N)}{\mu_F}) G(\tilde{u})\\
                                                    &+ \eta\rho||Ax^0 + B\omega^*||_2^2 + \frac{\eta}{\rho}||\lambda^0 - \lambda^*||_2^2
\end{aligned}
\end{equation}
Using (\ref{eq:strongly}) again and $x^0 = \tilde{x}, \omega^0 = \tilde{\omega}$, we have:
\begin{equation}
  \label{eq:r3}
  ||x^0 - x^*||_2^2 \leq \frac{2}{\mu_F}G(\tilde{u})
\end{equation}
Since $Ax^*+B\omega^* = 0$, we obtain:
\begin{equation}
  \begin{aligned}
    \label{eq:r1}
    \eta\rho||Ax^0 + B\omega^*||_2^2 &= \eta\rho||Ax^0 - Ax^*||_2^2\\
                                     &= \eta\rho|||x^0-x^*||_{A^{T}A}^2\\
                                     &\leq \eta\rho ||A^{T}A||\ ||x^0 - x^*||_2^2\\
                                     &\leq \frac{2\eta\rho||A^TA||}{\mu_F}G(\tilde{u})
  \end{aligned}
\end{equation}
Using the same technique we have the follwing bound:
\begin{equation}
  \begin{aligned}
    \label{eq:r2}
  \frac{\eta}{\rho}||\lambda^0 - \lambda^*||_2^2 &= \frac{\eta}{\rho}||\tilde{\lambda}^{s-1} - \lambda^*||_2^2\\
  &= \frac{\eta}{\rho}||-(A^{T})^{\dag}\nabla F(\tilde{x}) + (A^{T})^{\dag}\nabla F(x^*)||_2^2 \\
  &=\frac{\eta}{\rho}||\nabla F(\tilde{x}) - \nabla F(x^*)||_{A^{\dag}(A^{\dag})^{T}}^2\\
  &\leq \frac{\eta ||A^{\dag}(A^{\dag})^{T}||}{\rho}||\nabla F(\tilde{x}) - \nabla F(x^*)||_2^2\\
  &\leq \frac{2L_F\eta||A^{\dag}(A^{\dag})^{T}||}{\rho}(F(\tilde{x}) - F(x^*) - \langle\, \nabla F(x^*), \tilde{x} - x^*\rangle)\\
  &\leq \frac{2L_F\eta||A^{\dag}(A^{\dag})^{T}||}{\rho} G(\tilde{u})\\
  &=\frac{2L_F\eta}{\rho\sigma_{min}(AA^T)}G(\tilde{u})
  \end{aligned}
\end{equation}
where in the second inequality we use the Lipschitz property of $F$ induced from Proposition \ref{pro:F-lip}.
Substituting $(\ref{eq:r3})$, $(\ref{eq:r1})$, $(\ref{eq:r2})$ into $(\ref{eq:i1})$, we have:
\begin{equation}
  \begin{aligned}
    2\eta \mathbf{E}\sum\limits_{k=0}^{K-1}G(u^{k+1}) &\leq  \frac{2}{\mu_F}G(\tilde{u})  + (\frac{32\eta^2C_G^4L_f^2}{\mu_F N} +\frac{48\eta^2L_F^2+8\eta DC_GL_fL_G\sigma(N)}{\mu_F})\mathbf{E}(\sum\limits_{k=0}^{K-1}G(u^k))\\
    & + K(\frac{32\eta^2C_G^4L_f^2}{\mu_F N} +\frac{48\eta^2L_F^2+8\eta DC_GL_fL_G\sigma(N)}{\mu_F}) G(\tilde{u})\\
    &  + \frac{2\eta\rho||A^TA||}{\mu_F}G(\tilde{u}) + \frac{2L_F\eta}{\rho\sigma_{min}(AA^T)}G(\tilde{u})
  \end{aligned}
\end{equation}
Rearranging the terms we have:
\begin{equation}
  \begin{aligned}
    \label{eq:i2}
  &  2\eta \mathbf{E}(\sum\limits_{k=0}^{K-1}G(u^{k+1})) - (\frac{32\eta^2C_G^4L_f^2}{\mu_F N} +\frac{48\eta^2L_F^2+8\eta DC_GL_fL_G\sigma(N)}{\mu_F})\mathbf{E}(\sum\limits_{k=0}^{K-1}G(u^k)) \\
  &  \leq (\frac{2}{\mu_F} + K(\frac{32\eta^2C_G^4L_f^2}{\mu_F N} +\frac{48\eta^2L_F^2+8\eta DC_GL_fL_G\sigma(N)}{\mu_F}) + \frac{2\eta\rho||A^TA||}{\mu_F} + \frac{2L_F\eta}{\rho\sigma_{min}(AA^T)})G(\tilde{u})
  \end{aligned}
\end{equation}
Denote $\varrho = (\frac{32\eta^2C_G^4L_f^2}{\mu_F N} +\frac{48\eta^2L_F^2+8\eta DC_GL_fL_G\sigma(N)}{\mu_F})$, then the left side of (\ref{eq:i2}) equals:
\begin{equation}
  \begin{aligned}
    \sum\limits_{k=1}^{K}(2\eta - \varrho) E(G(u^k)) + \varrho E(G(u^K)) - \varrho G(u^0)
  \end{aligned}
\end{equation}
then we have:
\begin{equation}
  \begin{aligned}
    &\sum\limits_{k=1}^{K}(2\eta - \varrho) E(G(u^k))\\
    &\leq (\frac{2}{\mu_F} + (K+1)\varrho + \frac{2\eta\rho||A^TA||}{\mu_F} + \frac{2L_F\eta}{\rho\sigma_{min}(AA^T)} )G(\tilde{u})
  \end{aligned}
\end{equation}
Because $\tilde{u}^s = \frac{1}{K}\sum\limits_{k=1}^{K}u^{k}$ and function $G$ is convex, $\tilde{u} = \tilde{u}^{s-1}$ we have:
\begin{equation}
  \begin{aligned}
    &(2\eta - \varrho)K \mathbf{E}(G(\tilde{u}^s)) \\
    &\leq (\frac{2}{\mu_F} + (K+1)\varrho + \frac{2\eta\rho||A^TA||}{\mu_F} + \frac{2L_F\eta}{\rho\sigma_{min}(AA^T)})G(\tilde{u}^{s-1})
  \end{aligned}
\end{equation}
Denote $\gamma_1 = (2\eta - \varrho)K$, $\gamma_2 = \frac{2}{\mu_F} + (K+1)\varrho + \frac{2\eta\rho||A^TA||}{\mu_F} + \frac{2L_F\eta}{\rho\sigma_{min}(AA^T)}$, we have:
\begin{equation}
\gamma_1\mathbf{E}[G(\tilde{u}^s)] \leq \gamma_2 G(\tilde{u}^{s-1})
\end{equation}
\QEDB
\subsection{Proof of Theorem \ref{thm:2}}
\label{sec:proof of thm-2}
We analyze the convergence rate in Theorem \ref{thm:2} in this section. Denote $\mathbb{I}_k = \{i_k,j_k\}$.
\begin{lemma}  For any $\lambda \in R^{p}$, we have:
\begin{equation}
  \begin{aligned}
    \label{eq:lambda_2}
    -\langle\, Ax^{k+1}+B\omega^{k+1}, \lambda^{k+1} - \lambda^* - \lambda\rangle = \frac{1}{2\rho}(||\lambda^k - \lambda^* - \lambda||_2^2 - ||\lambda^{k+1} - \lambda^* - \lambda||_2^2 - ||\lambda^k - \lambda^{k+1}||_2^2)
  \end{aligned}
\end{equation}
\end{lemma}
\noindent
\emph{Proof.}
The proof is similar to Lemma $\ref{lemma:lambda}$, so we omit here.\QEDB
\\
\\
Similar to equation $(\ref{eq:gra})$, the gradient descent fomulation for $x$ update in Algorithm \ref{Algorithm-2} is:
\begin{equation}
  \begin{aligned}
    \label{eq:gra_2}
    x^{k+1} = x^k - \eta_s G_k^{-1}\mu_{i_k}^k\\
    \mu_{i_k}^{k} = \nabla \hat{F}_{i_k}(x^k) + A^{T}\lambda^{k+1}
  \end{aligned}
\end{equation}
\\
\begin{lemma}
  \label{lemma:7}
\begin{equation}
  \begin{aligned}
    \label{eq:lemma7}
  &-2\eta_s \langle\,\mu_{i_k}^{k}, x^k - x^* \rangle + \eta_s^2||\mu_{i_k}^k||_{G_k^{-1}}^2 \leq \\
  &-2\eta_s(F(x^{k+1}) - F(x^*)) - 2\eta_s\langle\,\nabla\hat{F}_{i_k}(x^k)-\nabla F(x^k), x^{k+1} - x^* \rangle + 2\eta_s \langle\, A^{T}\lambda^{k+1}, x^* - x^{k+1}\rangle
\end{aligned}
\end{equation}
\end{lemma}
\noindent
\emph{Proof.}
Using the same technique as (\ref{eq:eta}) in Lemma \ref{lemma:2}, we can get the following  corresponding inequality:
\begin{equation}
  \begin{aligned}
    &F(x^*) +  \langle\, A^{T}\lambda^{k+1}, x^{*} - x^{k+1}\rangle \\
    & \geq F(x^{k+1}) + \langle\, \nabla F(x^k) - \nabla\hat{F}_{i_k}(x^k) , x^{*} - x^{k+1}\rangle + \langle\, \mu_{i_k}^k, x^{*} -x^k\rangle \\
    &+ ||\mu_{i_k}^k||_{\eta_s G_k^{-1} - \frac{1}{2}L_F\eta_s^2 G_k^{-1}G_k^{-1}}^2
  \end{aligned}
\end{equation}
Next we prove that:
\begin{equation}
  \begin{aligned}
    ||\mu_{i_k}^k||_{\eta_s G_k^{-1} - \frac{1}{2}L_F\eta_s^2 G_k^{-1}G_k^{-1}}^2 \geq ||\mu_{i_k}^k||_{\frac{1}{2}\eta_sG_k^{-1}}^2
  \end{aligned}
\end{equation}
By the definition of $G$-norm, we only need to prove:
\begin{equation}
  \eta_s G_k^{-1} - \frac{1}{2}L_F\eta_s^2 G_k^{-1}G_k^{-1} \succeq \frac{1}{2}\eta_sG_k^{-1}
\end{equation}
it suffices to prove:
\begin{equation}
  L_F\eta_sG_k^{-1} \preceq I
\end{equation}
and it is true by the definition of $\eta_s$ and $G_k$. Therefore we have:
\begin{equation}
  \begin{aligned}
    &F(x^*) +  \langle\, A^{T}\lambda^{k+1}, x^{*} - x^{k+1}\rangle \\
    & \geq F(x^{k+1}) + \langle\, \nabla F(x^k) - \nabla\hat{F}_{i_k}(x^k) , x^{*} - x^{k+1}\rangle + \langle\, \mu_{i_k}^k, x^{*} -x^k\rangle \\
    & + \frac{1}{2}\eta_s||\mu_{i_k}^k||_{G_k^{-1}}^2
  \end{aligned}
\end{equation}
Multiplying two sides by $2\eta_s$ and rearranging terms, we have (\ref{eq:lemma7}).
\QEDB
\\
\\
\begin{lemma}
  \label{lemma:8}
\begin{equation}
  \begin{aligned}
    \label{eq:lemma8}
  \mathbf{E}  ||x^{k+1} - x^*||_{G_k}^2 \leq & \mathbf{E}||x^k - x^*||_{G_k}^2 -2\eta_s\mathbf{E}(F(x^{k+1}) - F(x^*)) - 2\eta_s\mathbf{E}\langle\,\nabla\hat{F}_{i_k}(x^k)-\nabla F(x^k), x^{k+1} - x^* \rangle \\
  &+ 2\eta_s \mathbf{E}\langle\, A^{T}\lambda^{k+1}, x^* - x^{k+1}\rangle
  \end{aligned}
\end{equation}
\end{lemma}
\noindent
\emph{Proof.} From (\ref{eq:gra_2}) we know that
\begin{equation}
  x^{k+1} - x^* = x^k - x^* - \eta_s G_k^{-1}\mu_{i_k}^k
\end{equation}
therefore by the definition of $G$-norm we have:
\begin{equation}
  \label{eq:i6}
  ||x^{k+1} - x^*||_{G_k}^2 = ||x^k - x^*||_{G_k}^2 - 2\eta_s\langle\, x^k - x^*, \mu_{i_k}^k\rangle + \eta_s^2||\mu_{i_k}^k||_{G_k^{-1}}^2
\end{equation}
By Lemma \ref{lemma:7} we obtain
\begin{equation}
  \begin{aligned}
    ||x^{k+1} - x^*||_{G_k}^2 & \leq ||x^k - x^*||_{G_k}^2 -2\eta_s(F(x^{k+1}) - F(x^*)) - 2\eta_s\langle\,\nabla\hat{F}_{i_k}(x^k)-\nabla F(x^k), x^{k+1} - x^* \rangle \\
    & + 2\eta_s \langle\, A^{T}\lambda^{k+1}, x^* - x^{k+1}\rangle
  \end{aligned}
\end{equation}
Taking expectation of $\mathbb{I}_k$ in the current step $s$,  we have Lemma \ref{lemma:8}.\QEDB
\\
\\
\noindent
\emph{Proof \ of \ Theorem \ref{thm:2}.} \
Taking expectation of $\mathbb{I}_k$ on (\ref{eq:ref-1}) in the current step $s$, we have
\begin{equation}
\begin{aligned}
  \label{eq:i7}
  -2\eta_s  \mathbf{E}\langle\, \nabla F(x^k) - \nabla\hat{F}_{i_k}(x^k) , x^{*} - x^{k+1}\rangle \leq 2\eta_s^2\underbrace{\mathbf{E}||\nabla\hat{F}_{i_k}(x^k) - \nabla F(x^k) ||_2^2}_{T_7} - 2\eta_s \mathbf{E}\langle\, \nabla\hat{F}_{i_k}(x^k) -\nabla F(x^k) , \bar{x} - x^{*}\rangle
\end{aligned}
\end{equation}
Note that $\mathbf{E}\langle\, \nabla\hat{F}_{i_k}(x^k) -\nabla F(x^k) , \bar{x} - x^{*}\rangle = 0$.
Now we bound $T_7$.\\
\begin{equation}
  \begin{aligned}
T_7 = &\mathbf{E}||\nabla\hat{F}_{i_k}(x^k) - \nabla F(x^k)||_2^2 \\
=& \mathbf{E}||(\partial g_{j_k}(x^k))^{T}\nabla f_{i_k}(g(x^k)) - (\partial g_{j_k}(\tilde{x}))^{T}\nabla  f_{i_k}(g(\tilde{x})) + (\partial g(\tilde{x}))^{T}\nabla f(g(\tilde{x})) \\
&- (\partial g(x^k))^{T}\nabla f(g(x^k))||_2^2\\
= & \mathbf{E}|| \ [(\partial g_{j_k}(x^k))^{T}\nabla f_{i_k}(g(x^k)) - (\partial g_{j_k}(\tilde{x}))^{T}\nabla  f_{i_k}(g(\tilde{x}))] \\
&- [ (\partial g(x^k))^{T}\nabla f(g(x^k)) - (\partial g(\tilde{x}))^{T}\nabla f(g(\tilde{x}))] \ ||_2^2\\
\leq & \mathbf{E} || (\partial g_{j_k}(x^k))^{T}\nabla f_{i_k}(g(x^k)) - (\partial g_{j_k}(\tilde{x}))^{T}\nabla  f_{i_k}(g(\tilde{x}))||_2^2 \quad \quad (E||x-Ex||_2^2 \leq E||x||_2^2)\\
= & \mathbf{E} || (\partial g_{j_k}(x^k))^{T}\nabla f_{i_k}(g(x^k)) - (\partial g_{j_k}(x^*))^{T}\nabla f_{i_k}(g(x^*)) \\
& + (\partial g_{j_k}(x^*))^{T}\nabla f_{i_k}(g(x^*))- (\partial g_{j_k}(\tilde{x}))^{T}\nabla  f_{i_k}(g(\tilde{x}))||_2^2\\
\leq & 2 \mathbf{E} || (\partial g_{j_k}(x^k))^{T}\nabla f_{i_k}(g(x^k)) - (\partial g_{j_k}(x^*))^{T}\nabla f_{i_k}(g(x^*))||_2^2 \\
&+ 2\mathbf{E} ||(\partial g_{j_k}(x^*))^{T}\nabla f_{i_k}(g(x^*))- (\partial g_{j_k}(\tilde{x}))^{T}\nabla  f_{i_k}(g(\tilde{x}))||_2^2\\
\leq & 2L_F^2\mathbf{E}||x^k - x^*||_2^2 + 2L_F^2||\tilde{x} - x^*||_2^2 \quad \quad (Assumption \ \ref{assump:7})
\end{aligned}
\end{equation}
therefore we have:
\begin{equation}
  \begin{aligned}
  & -2\eta_s  \mathbf{E}\langle\, \nabla F(x^k) - \nabla\hat{F}_{i_k}(x^k) , x^{*} - x^{k+1}\rangle \\
  & \leq  4L_F^2\eta_s^2\mathbf{E}||x^k - x^*||_2^2 + 4L_F^2\eta_s^2||\tilde{x} - x^*||_2^2 \\
  & \leq 8L_F^2\eta_s^2D^2
  \end{aligned}
\end{equation}
Then equation (\ref{eq:lemma8}) becomes:
\begin{equation}
  \begin{aligned}
    \mathbf{E}  ||x^{k+1} - x^*||_{G_k}^2 \leq & \mathbf{E}||x^k - x^*||_{G_k}^2 -2\eta_s\mathbf{E}(F(x^{k+1}) - F(x^*)) + 8L_F^2\eta_s^2D^2 + 2\eta_s \mathbf{E}\langle\, A^{T}\lambda^{k+1}, x^* - x^{k+1}\rangle
  \end{aligned}
\end{equation}
Rearranging terms and using $G_{k} \succeq G_{k+1}$, we have:
\begin{equation}
  \begin{aligned}
    2\eta_s\mathbf{E}(F(x^{k+1}) - F(x^*) - \langle\, A^T\lambda^{k+1}, x^* - x^{k+1}\rangle) & \leq \mathbf{E}||x^k - x^*||_{G_k}^2 - \mathbf{E}||x^{k+1} - x^*||_{G_k}^2 + 8L_F^2\eta_s^2D^2\\
    & \leq \mathbf{E}||x^k - x^*||_{G_k}^2 - \mathbf{E}||x^{k+1} - x^*||_{G_{k+1}}^2 + 8L_F^2\eta_s^2D^2
  \end{aligned}
\end{equation}
Using (\ref{eq:F_left}) we have:
\begin{equation}
  \begin{aligned}
    \label{eq:x_2}
    & 2\eta_s\mathbf{E}(F(x^{k+1}) - F(x^*) - \langle\,\nabla F(x^*), x^{k+1} - x^*\rangle + \langle\, \lambda^{k+1} - \lambda^*, Ax^{k+1} - Ax^*\rangle)\\
    & \leq \mathbf{E}||x^k - x^*||_{G_k}^2 - \mathbf{E}||x^{k+1} - x^*||_{G_{k+1}}^2 + 8L_F^2\eta_s^2D^2
  \end{aligned}
\end{equation}
Calculating $(\ref{eq:x_2}) + (\ref{eq:omega})\times 2\eta_s + (\ref{eq:lambda_2}) \times 2\eta_s$ and using $\lambda^{k+1} - \lambda^k = \rho(Ax^{k+1} + B\omega^{k+1})$ we have:
\begin{equation}
  \begin{aligned}
    & 2\eta_s\mathbf{E}(F(x^{k+1}) - F(x^*) - \langle\, \nabla F(x^*), x^{k+1} - x^*\rangle + R(\omega^{k+1}) - R(\omega^*) - \langle\, \tilde{\nabla}R(\omega^*), \omega^{k+1} - \omega^*\rangle + \langle\, \lambda, Ax^{k+1}+B\omega^{k+1}\rangle)\\
    & \leq \mathbf{E}||x^{k} - x^*||_{G_k}^2 - \mathbf{E}||x^{k+1} - x^*||_{G_{k+1}}^2 + 8L_F^2\eta_s^2D^2\\
    &+\rho\eta_s(\mathbf{E}||Ax^k + B\omega^*||_2^2 - \mathbf{E}||Ax^{k+1} + B\omega^{*}||_2^2)\\
    &+ \frac{\eta_s}{\rho}(\mathbf{E}||\lambda^k - \lambda^* - \lambda||_2^2 - \mathbf{E}||\lambda^{k+1} - \lambda^* - \lambda||_2^2)
  \end{aligned}
\end{equation}
Summing from $k=0$ to $K-1$, using the definition of $\tilde{x}^s,\tilde{\omega}^s$ and the convexity of $F$ and $R$, we have:
\begin{equation}
  \begin{aligned}
    \label{eq:i3}
    & 2\eta_sK\mathbf{E}(F(\tilde{x}^s) - F(x^*) - \langle\, \nabla F(x^*), \tilde{x}^s - x^*\rangle + R(\tilde{\omega}^s) - R(\omega^*) - \langle\, \tilde{\nabla}R(\omega^*), \tilde{\omega}^s - \omega^*\rangle + \langle\, \lambda, A\tilde{x}^s+B\tilde{\omega}^s \rangle)\\
    & \leq \mathbf{E}(||x^0 - x^*||_{G_0}^2 - ||x^K - x^*||_{G_K}^2) + 8L_F^2\eta_s^2D^2K\\
    & + \rho\eta_s\mathbf{E}(||Ax^0 + B\omega^*||_2^2 - ||Ax^K + B\omega^*||_2^2) + \frac{\eta_s}{\rho}\mathbf{E}(||\lambda^0 - \lambda^* - \lambda||_2^2 - ||\lambda^K - \lambda^*-\lambda||_2^2)
  \end{aligned}
  \end{equation}
Recall the definitions: $x^0 = \hat{x}^{s-1}, x^K = \hat{x}^s, \lambda^0 = \hat{\lambda}^{s-1}, \lambda^K = \hat{\lambda}^s, G_0 = \hat{G}^{s-1} = \frac{1}{s}I, G_K = \hat{G}_s = \frac{1}{s+1}I, \eta_s = \frac{1}{(s+1)L_F}$, then taking expectation on all steps $s$ and (\ref{eq:i3}) becomes:
\begin{equation}
  \begin{aligned}
&  2\eta_sK\mathbf{E}(F(\tilde{x}^s) - F(x^*) - \langle\, \nabla F(x^*), \tilde{x}^s - x^*\rangle + R(\tilde{\omega}^s) - R(\omega^*) - \langle\, \tilde{\nabla}R(\omega^*), \tilde{\omega}^s - \omega^*\rangle + \langle\, \lambda, A\tilde{x}^s+B\tilde{\omega}^s \rangle)\\
&  \leq \mathbf{E}(||\hat{x}^{s-1} - x^*||_{\hat{G}^{s-1}}^2 - ||\hat{x}^s - x^*||_{\hat{G}^s}^2) + 8L_F^2\eta_s^2D^2K\\
  & + \rho\eta_s \mathbf{E}(||A\hat{x}^{s-1} + B\omega^*||_2^2 - ||A\hat{x}^s + B\omega^*||_2^2) + \frac{\eta_s}{\rho}\mathbf{E}(||\hat{\lambda}^{s-1} - \lambda^* - \lambda||_2^2 - ||\hat{\lambda}^{s} - \lambda^*-\lambda||_2^2)
  \end{aligned}
\end{equation}
Dividing both sides of $\eta_s$ and summing over $s = 1,...,S$, we have:
\begin{equation}
\begin{aligned}
  \label{eq:i4}
  & 2KS\mathbf{E}( F(\bar{x}) - F(x^*) - \langle\, \nabla F(x^*), \bar{x} - x^*\rangle + R(\bar{\omega}) - R(\omega^*) - \langle\, \tilde{\nabla}R(\omega^*), \bar{\omega}-\omega^*\rangle + \langle\, \lambda, A\bar{x} + B\bar{\omega}\rangle)\\
  & \leq \underbrace{\sum\limits_{s=1}^{S}\mathbf{E}(\frac{1}{\eta_s}||\hat{x}^{s-1} - x^*||_{\hat{G}^{s-1}}^2 - \frac{1}{\eta_s}||\hat{x}^{s} - x^*||_{\hat{G}^{s}}^2)}_{T_8} + \underbrace{\sum\limits_{s=1}^{S}(8L_F^2\eta_sD^2K)}_{T_9} +\rho||A\hat{x}^0 + B\omega^*||_2^2 + \underbrace{\frac{1}{\rho}||\hat{\lambda}^0 - \lambda^* - \lambda||_2^2}_{T_{10}}
\end{aligned}
\end{equation}
where we use $\bar{x} = \frac{1}{S}\sum\limits_{s=1}^S\tilde{x}^s$, $\bar{\omega} = \frac{1}{S}\sum\limits_{s=1}^S\tilde{\omega}^s$ and the convexity of $F,R$ again.\\
Letting $\lambda = \Lambda\frac{A\bar{x}+B\bar{\omega}}{||A\bar{x}+B\bar{\omega}||}$, $\Lambda > 0$, then the left side of (\ref{eq:i4}) equals:
\begin{equation}
  2KS\mathbf{E}( F(\bar{x}) - F(x^*) - \langle\, \nabla F(x^*), \bar{x} - x^*\rangle + R(\bar{\omega}) - R(\omega^*) - \langle\, \tilde{\nabla}R(\omega^*), \bar{\omega}-\omega^*\rangle +\Lambda||A\bar{x} + B\bar{\omega}||)
\end{equation}
Now we bound $T_8$ by the definition of $G$-norm and $\hat{G}^{s}$, $\eta_s$.
\begin{equation}
\begin{aligned}
  T_8 = &\sum\limits_{s=1}^{S} \mathbf{E}(\frac{1}{\eta_s}||\hat{x}^{s-1} - x^*||_{\hat{G}^{s-1}}^2 - \frac{1}{\eta_s}||\hat{x}^{s} - x^*||_{\hat{G}^{s}}^2)\\
  &=\mathbf{E} \sum\limits_{s=1}^{S}(||\hat{x}^{s-1} - x^*||_{\frac{\hat{G}^{s-1}}{\eta_s}}^2 - ||\hat{x}^{s} - x^*||_{\frac{\hat{G}^{s}}{\eta_s}}^2)\\
  &= \mathbf{E}\sum\limits_{s=1}^{S}(||\hat{x}^{s-1} - x^*||_{\frac{L_F(s+1)}{s}}^2 - ||\hat{x}^{s} - x^*||_{L_F}^2)\\
  &=L_F\mathbf{E}\sum\limits_{s=1}^{S} (\frac{1}{s} ||\hat{x}^{s-1} - x^*||_2^2 + ||\hat{x}^{s-1} - x^*||_2^2 - ||\hat{x}^{s} - x^*||_2^2)\\
  &\leq L_FD^2\sum\limits_{s=1}^{S}\frac{1}{s} + L_F\mathbf{E}\sum\limits_{s=1}^{S}(||\hat{x}^{s-1} - x^*||_2^2 - ||\hat{x}^{s} - x^*||_2^2)\\
  &\leq L_FD^2\log{S} + L_FD^2
\end{aligned}
\end{equation}
And $T_9 = \sum\limits_{s=1}^{S}(8L_F^2\eta_sD^2K) \leq 8L_FD^2K\log(S+1)$. Using the definition of $\lambda$ we have:
\begin{equation}
  T_{10} = \frac{1}{\rho}||\hat{\lambda}^0 - \lambda^* - \lambda||_2^2 \leq \frac{2}{\rho}||\hat{\lambda}^0 - \lambda^*||_2^2 + \frac{2}{\rho}||\lambda||_2^2 = \frac{2}{\rho}||\hat{\lambda}^0 - \lambda^*||_2^2 + \frac{2}{\rho}\Lambda^2
\end{equation}
Substituting the bound of $T_8, T_9, T_{10}$ back into (\ref{eq:i4}) and bound $\rho||A\hat{x}^0 + B\omega^*||_2^2$ as the first inequality in (\ref{eq:r1}), dividing both sides by $2KS$ we have:
\begin{equation}
\begin{aligned}
  & \mathbf{E}( F(\bar{x}) - F(x^*) - \langle\, \nabla F(x^*), \bar{x} - x^*\rangle + R(\bar{\omega}) - R(\omega^*) - \langle\, \tilde{\nabla}R(\omega^*), \bar{\omega}-\omega^*\rangle + \Lambda||A\bar{x} + B\bar{\omega}||)\\
  & \leq \frac{4L_FD^2\log(S+1)}{S} + \frac{L_FD^2\log{S}}{2KS} +\frac{L_FD^2 + \rho D^2||A^TA|| + \frac{2}{\rho} ||\hat{\lambda}^0 - \lambda^*||_2^2 + \frac{2}{\rho}\Lambda^2}{2KS}
\end{aligned}
\end{equation}
\QEDB
\subsection{Proof of Theorem \ref{thm:3}}
Using the same definition of $\mathbb{I}_k$ as in section \ref{sec:proof of thm-2}, we prove the convergence rate in Theorem \ref{thm:3}.
\begin{lemma} If $B$ is invertible, lettting $s^{k+1} \in \partial R(\omega^{k+1})$, we have: \label{lemma:9}
\begin{equation}
  \label{eq:inv-B}
\lambda^{k+1} - \rho(Ax^{k+1} - Ax^k)  = -(B^{T})^{-1}s^{k+1}
\end{equation}
\end{lemma}
\noindent
\emph{Proof.}\
According to (\ref{eq:opt-omega}) and using $\lambda^{k+1} = \lambda^k + \rho(Ax^{k+1} + B\omega^{k+1})$, we have:
\begin{equation}
  \begin{aligned}
  0 &=  s^{k+1} + B^{T}\lambda^{k} + \rho B^{T}(Ax^{k} + B\omega^{k+1})\\
    &=  s^{k+1} + B^{T}\lambda^{k+1} + \rho B^{T}(-Ax^{k+1} + Ax^k)
\end{aligned}
\end{equation}
\begin{equation}
  \begin{aligned}
    \therefore  B^{T}\lambda^{k+1} - \rho B^T(Ax^{k+1} - Ax^k) = -s^{k+1}
  \end{aligned}
\end{equation}
mutiplying two sides by $(B^T)^{-1}$, we obtain Lemma \ref{lemma:9}.\QEDB
\\
\\
\noindent
\begin{lemma}
  \label{lemma:10}
\begin{equation}
  ||x^{k+1} - x^k||_2  \leq \frac{C_1}{\sqrt{s}}
\end{equation}
where $C_1$ is some positive constant.
\end{lemma}
\noindent
\emph{Proof.}
According to (\ref{eq:gra_2}), we have:
\begin{equation}
  \begin{aligned}
    x^{k+1} - x^k  &= -\eta_sG_k^{-1}(\nabla\hat{F}_{i_k}(x^k) + A^T\lambda^{k+1}) \\
    &= -\eta_sG_k^{-1}\nabla\hat{F}_{i_k}(x^k) - \eta_sG_k^{-1}A^T(\lambda^{k+1} - \rho(Ax^{k+1} - Ax^k)) - \rho \eta_sG_k^{-1}A^T(Ax^{k+1} - Ax^k)\\
    &=-\eta_sG_k^{-1}\nabla\hat{F}_{i_k}(x^k) + \eta_sG_k^{-1}A^T(B^T)^{-1}s^{k+1} - \rho \eta_sG_k^{-1}A^T(Ax^{k+1} - Ax^k)
  \end{aligned}
\end{equation}
where we use (\ref{eq:inv-B}) in the third equality. Multiplying two sides by $G_k$ we have:
\begin{equation}
  \begin{aligned}
    G_k(x^{k+1} - x^k) = -\eta_s(\nabla\hat{F}_{i_k}(x^k) - A^T(B^T)^{-1}s^{k+1}) - \rho \eta_sA^T(Ax^{k+1} - Ax^k)
  \end{aligned}
  \end{equation}
  \begin{equation}
    \begin{aligned}
      \therefore ||G_k(x^{k+1} - x^k)||_2^2 &\leq 2\eta_s^2 ||\nabla\hat{F}_{i_k}(x^k) - A^T(B^T)^{-1}s^{k+1}||_2^2 + 2\rho^2\eta_s^2||A^TA(x^{k+1} - x^k)||_2^2\\
      &\leq \eta_s^2 C_1 + \eta_s^2 C_2||x^{k+1}-x^k||_2^2
    \end{aligned}
    \end{equation}
where $C_1 > 1$ and we use Assumption \ref{assump:8} in the second inequality, so that $2||\nabla\hat{F}_{i_k}(x^k) - A^T(B^T)^{-1}s^{k+1}||_2^2 \leq C_1$ and $ 2\rho^2||A^TAA^TA||\leq C_2$.  Because $\frac{1}{\sqrt{s+1}}I \leq G_k \leq \frac{1}{\sqrt{s}}I$ and $\eta_s = \frac{1}{s+1}$, we obtain :
\begin{equation}
  \begin{aligned}
    \frac{1}{s+1}||x^{k+1} - x^k||_2^2 \leq \frac{1}{(s+1)^2} C_1 + \frac{1}{(s+1)^2}C_2||x^{k+1} - x^k||_2^2
  \end{aligned}
  \end{equation}
  Multiplying two sides by $(s+1)^2$ and choosing $\rho$ such that $1-C_2 > 0$, after rearranging terms, we have
  \begin{equation}
    \begin{aligned}
    ||x^{k+1} - x^k||_2 \leq \frac{\sqrt{C_1}}{\sqrt{s}} \leq \frac{C_1}{\sqrt{s}}
    \end{aligned}
    \end{equation}
\QEDB\\
\\
\noindent
\emph{Proof \ of \ Theorem \ref{thm:3}}. Using the notation of (\ref{eq:gra_2}), we have:
\begin{equation}
  \begin{aligned}
    F(x^*) & \geq F(x^k) + \langle\, \nabla F(x^k), x^* - x^k\rangle\\
    & = F(x^k) + \langle\, \nabla F(x^k), x^* - x^{k+1}\rangle + \langle\, \nabla F(x^k), x^{k+1} - x^k\rangle
  \end{aligned}
\end{equation}
\begin{equation}
  \begin{aligned}
    \because \  & \langle\, \nabla F(x^k), x^* - x^{k+1}\rangle + \langle\, A^T\lambda^{k+1}, x^* - x^{k+1}\rangle\\
             & = \langle\, \nabla F(x^k), x^* - x^{k+1}\rangle + \langle\, \mu_{i_k}^k - \nabla\hat{F}_{i_k}(x^k), x^* - x^{k+1}\rangle\\
             & = \langle\, \mu_{i_k}^k, x^* - x^{k+1}\rangle + \langle\, \nabla\hat{F}_{i_k}(x^k) - \nabla F_(x^k), x^{k+1} - x^{*}\rangle \\
             &= \langle\, \mu_{i_k}^k, x^* - x^{k}\rangle + \langle\, \mu_{i_k}^k, x^k - x^{k+1}\rangle + \langle\, \nabla\hat{F}_{i_k}(x^k) - \nabla F(x^k), x^{k+1} - x^{*}\rangle\\
             & = \langle\, \mu_{i_k}^k, x^* - x^{k}\rangle + \langle\, \mu_{i_k}^k, \eta_sG_k^{-1}\mu_{i_k}^k\rangle + \langle\, \nabla\hat{F}_{i_k}(x^k) - \nabla F(x^k), x^{k+1} - x^{*}\rangle
  \end{aligned}
\end{equation}
Therefore:
\begin{equation}
  \begin{aligned}
    \label{eq:i5}
    & F(x^*) + \langle\, A^T\lambda^{k+1}, x^* - x^{k+1}\rangle \\
    & \geq F(x^k) + \langle\, \mu_{i_k}^k, x^* - x^{k}\rangle + \eta_s||\mu_{i_k}^k||_{G_k^{-1}}^2 + \langle\, \nabla\hat{F}_{i_k}(x^k) - \nabla F(x^k), x^{k+1} - x^{*}\rangle + \langle\, \nabla F(x^k), x^{k+1} - x^k\rangle\\
    & \geq F(x^k) + \langle\, \mu_{i_k}^k, x^* - x^{k}\rangle + \frac{\eta_s}{2}||\mu_{i_k}^k||_{G_k^{-1}}^2 + \langle\, \nabla\hat{F}_{i_k}(x^k) - \nabla F(x^k), x^{k+1} - x^{*}\rangle + \langle\, \nabla F(x^k), x^{k+1} - x^k\rangle
  \end{aligned}
\end{equation}
Multiplying both sides of (\ref{eq:i5}) by $2\eta_s$ and rearranging terms, we obtain:
\begin{equation}
  \begin{aligned}
    & -2\eta_s \langle\, \mu_{i_k}^k, x^k - x^{*}\rangle + \eta_s^2||\mu_{i_k}^k||_{G_k^{-1}}^2\\
    & \leq -2\eta_s(F(x^k) - F(x^*)) + 2\eta_s\langle\, A^T\lambda^{k+1}, x^* - x^{k+1}\rangle -2\eta_s\langle\, \nabla\hat{F}_{i_k}(x^k) - \nabla F(x^k), x^{k+1} - x^{*}\rangle \\
    &- 2\eta_s\langle\, \nabla F(x^k), x^{k+1} - x^k\rangle
  \end{aligned}
\end{equation}
then the equation (\ref{eq:i6}) is bounded by:
\begin{equation}
  \begin{aligned}
    \label{eq:i8}
    ||x^{k+1} - x^*||_{G_k}^2 \leq & ||x^k - x^*||_{G_k}^2 -2\eta_s(F(x^k) - F(x^*)) + 2\eta_s\langle\, A^T\lambda^{k+1}, x^* - x^{k+1}\rangle \\
    & \underbrace{-2\eta_s\langle\, \nabla\hat{F}_{i_k}(x^k) - \nabla F(x^k), x^{k+1} - x^{*}\rangle}_{T_{11}} \underbrace{- 2\eta_s\langle\, \nabla F(x^k), x^{k+1} - x^k\rangle}_{T_{12}}
  \end{aligned}
\end{equation}
 For $T_{12}$ we have:
\begin{equation}
  \begin{aligned}
    \label{eq:i9}
    T_{12} &= 2\eta_s\langle\, \nabla F(x^k), x^{k} - x^{k+1}\rangle\\
    &\leq 2\eta_s||\nabla F(x^k)||_2 ||x^{k+1} - x^k||_2\\
    &\leq 2\eta_sC_F||x^{k+1} - x^k||_2\\
    &\leq \frac{C_1C_F}{(s+1)\sqrt{s}}
  \end{aligned}
\end{equation}
In the second inequality we use the assumption that $||\nabla F(x)|| \leq C_F$ for all $x$. In the third inequality, we use Lemma \ref{lemma:10} \ (we hide constant 2 in constant $C_1$).
 Taking expectation of $\mathbb{I}_k$ on (\ref{eq:i8}) in the current step $s$ and using (\ref{eq:i7}, \ref{eq:i9}), after rearranging terms we have
\begin{equation}
  \begin{aligned}
    \label{ea:i10}
    & \mathbf{E}(2\eta_s(F(x^k) - F(x^*)) - 2\eta_s\langle\, A^T\lambda^{k+1}, x^* - x^{k+1}\rangle)  \\
    & \leq \mathbf{E}||x^k - x^*||_{G_k}^2 - \mathbf{E}||x^{k+1} - x^*||_{G_k}^2 + 2\eta_s^2\underbrace{\mathbf{E}||\nabla\hat{F}_{i_k}(x^k) - \nabla F(x^k) ||_2^2}_{T_{13}} + \frac{C_1C_F}{(s+1)\sqrt{s}}
  \end{aligned}
\end{equation}
where we use $\mathbf{E}\langle\, \nabla\hat{F}_{i_k}(x^k) -\nabla F(x^k) , \bar{x} - x^{*}\rangle = 0$. Moreover, under Assumption \ref{assump:8}, we obtain the bound for $T_{13}$:
\begin{equation}
  \begin{aligned}
    T_{13}&= \mathbf{E}||\nabla\hat{F}_{i_k}(x^k) - \nabla F(x^k) ||_2^2 \\
    & = \mathbf{E}||\partial g_{j_k}(x^k)\nabla f_{i_k}(g(x^k))  - \partial g_{j_k}(\tilde{x})\nabla f_{i_k}(g(\tilde{x}))  + \partial g(\tilde{x})\nabla f(g(\tilde{x})) - \partial g(x^k)\nabla f(g(x^k)) ||_2^2\\
    & \leq \mathbf{E} ||\partial g_{j_k}(x^k)\nabla f_{i_k}(g(x^k))  - \partial g_{j_k}(\tilde{x})\nabla f_{i_k}(g(\tilde{x}))||_2^2 \quad \quad (\mathbf{E}||x-Ex||_2^2 \leq \mathbf{E}||x||_2^2)\\
    &\leq C_3
  \end{aligned}
\end{equation}
where $C_3$ is a positive constant. Therefore equation (\ref{ea:i10}) becomes:
\begin{equation}
  \begin{aligned}
    \label{eq:i11}
    & \mathbf{E}(2\eta_s(F(x^k) - F(x^*)) - 2\eta_s\langle\, A^T\lambda^{k+1}, x^* - x^{k+1}\rangle)  \\
    & \leq \mathbf{E}||x^k - x^*||_{G_k}^2 - \mathbf{E}||x^{k+1} - x^*||_{G_k}^2 + 2\eta_s^2C_3 + \frac{C_1C_F}{(s+1)\sqrt{s}}
  \end{aligned}
\end{equation}
We then bound the left side of (\ref{eq:i11}):
\begin{equation}
  \begin{aligned}
    & \mathbf{E}(2\eta_s(F(x^k) - F(x^*)) - 2\eta_s\langle\, A^T\lambda^{k+1}, x^* - x^{k+1}\rangle) \\
    & = 2\eta_s \mathbf{E}(F(x^k) - F(x^*) - \langle\, \nabla F(x^*), x^k - x^*\rangle + \langle\, \nabla F(x^*), x^k - x^*\rangle + \langle\, A^T\lambda^{k+1}, x^{k+1} - x^{*}\rangle)\\
    &= 2\eta_s \mathbf{E}(F(x^k) - F(x^*) - \langle\, \nabla F(x^*), x^k - x^*\rangle + \langle\, -A^T\lambda^*, x^k - x^*\rangle + \langle\, A^T\lambda^{k+1}, x^{k+1} - x^{*}\rangle)\\
    &= 2\eta_s \mathbf{E}(F(x^k) - F(x^*) - \langle\, \nabla F(x^*), x^k - x^*\rangle + \langle\, \lambda^{k+1} - \lambda^*, Ax^{k+1} - Ax^*\rangle) + 2\eta_s \mathbf{E}\langle\, A^T\lambda^{*}, x^{k+1} - x^{k}\rangle\\
  \end{aligned}
\end{equation}
Putting back into (\ref{eq:i11}) we have:
\begin{equation}
  \begin{aligned}
    &2\eta_s \mathbf{E}(F(x^k) - F(x^*) - \langle\, \nabla F(x^*), x^k - x^*\rangle + \langle\, \lambda^{k+1} - \lambda^*, Ax^{k+1} - Ax^*\rangle)\\
    &\leq  \underbrace{-2\eta_s \mathbf{E}\langle\, A^T\lambda^{*}, x^{k+1} - x^{k}\rangle}_{T_{14}} + \mathbf{E}||x^k - x^*||_{G_k}^2 - \mathbf{E}||x^{k+1} - x^*||_{G_k}^2 + 2\eta_s^2C_3 + \frac{C_1C_F}{(s+1)\sqrt{s}}
  \end{aligned}
\end{equation}
We then bound $T_{14}$:
\begin{equation}
  \begin{aligned}
    T_{14} \leq 2\eta_s\mathbf{E}||A^T\lambda^*||_2||x^{k+1} - x^k||_2 \leq C_4\eta_s\mathbf{E}||x^{k+1} - x^{k}||_2 \leq \frac{C_1 C_4}{(s+1)\sqrt{s}}
  \end{aligned}
\end{equation}
where $2||A^T\lambda^*||_2\leq C_4$ and $C_4 > 0$.
\begin{equation}
  \begin{aligned}
    \label{eq:x-3}
    \therefore \ & 2\eta_s \mathbf{E}(F(x^k) - F(x^*) - \langle\, \nabla F(x^*), x^k - x^*\rangle + \langle\, \lambda^{k+1} - \lambda^*, Ax^{k+1} - Ax^*\rangle)\\
    & \leq  \mathbf{E}||x^k - x^*||_{G_k}^2 - \mathbf{E}||x^{k+1} - x^*||_{G_k}^2 + 2\eta_s^2C_3 + \frac{C_1C_F + C_1C_4}{(s+1)\sqrt{s}}
  \end{aligned}
\end{equation}
After calculating $(\ref{eq:x-3}) + (\ref{eq:omega})\times 2\eta_s + (\ref{eq:lambda_2}) \times 2\eta_s$, we have
\begin{equation}
\begin{aligned}
  & 2\eta_s \mathbf{E}(F(x^k) - F(x^*) - \langle\, \nabla F(x^*), x^k - x^*\rangle + R(\omega^{k+1}) - R(\omega^*) - \langle\, \tilde{\nabla}R(\omega^*), \omega^{k+1} - \omega^*\rangle + \langle\, \lambda, Ax^{k+1}+B\omega^{k+1})\\
  & \leq \mathbf{E}||x^k - x^*||_{G_k}^2 - \mathbf{E}||x^{k+1} - x^*||_{G_{k+1}}^2 +2\eta_s^2C_3 + \frac{C_1(C_4+C_F)}{(s+1)\sqrt{s}}\\
  & + \rho\eta_s\mathbf{E}(||Ax^k+B\omega^*||_2^2 - ||Ax^{k+1} + B\omega^*||_2^2) + \frac{\eta_s}{\rho}\mathbf{E}(||\lambda^k - \lambda^* - \lambda||_2^2 - ||\lambda^{k+1} - \lambda^* - \lambda||_2^2)
\end{aligned}
\end{equation}
Recall the definition of $z, \tilde{z}^s$,  summing from $k=0$ to $K-1$ we have:
\begin{equation}
  \begin{aligned}
    & 2\eta_s K \mathbf{E}(G(\tilde{z}^s)  + \langle\, \lambda, A\tilde{x}^s + B\tilde{\omega}^s\rangle) \\
    & \leq \mathbf{E} (||x^0 - x^*||_{G_0}^2 - ||x^K - x^*||_{G_K}^2) + 2\eta_s^2C_3K + K\frac{C_1(C_4+C_F)}{(s+1)\sqrt{s}}\\
    & + \rho\eta_s\mathbf{E}(||Ax^0 + B\omega^*||_2^2 - ||Ax^K + B\omega^*||_2^2) + \frac{\eta_s}{\rho}\mathbf{E}(||\lambda^0 - \lambda^* - \lambda||_2^2 - ||\lambda^K - \lambda^* - \lambda||_2^2)
  \end{aligned}
\end{equation}
Dividing both sides of $\eta_s$ and using the definition of $\hat{x}^{s-1}$, $\hat{x}^{s}$, $\hat{\lambda}^{s-1}$, $\hat{\lambda}^{s}$, if taking expectation over all steps $s$ we have:
\begin{equation}
  \begin{aligned}
    & 2 K \mathbf{E}(G(\tilde{z}^s)  + \langle\, \lambda, A\tilde{x}^s + B\tilde{\omega}^s\rangle)  \\
    & \leq \frac{1}{\eta_s}\mathbf{E} (||\hat{x}^{s-1} - x^*||_{\hat{G}^{s-1}}^2 - ||\hat{x}^s - x^*||_{\hat{G}^s}^2) + 2\eta_sC_3K + K\frac{ C_1(C_4+C_F)}{\sqrt{s}}\\
    & + \rho\mathbf{E}(||A\hat{x}^{s-1} + B\omega^*||_2^2 - ||A\hat{x}^{s}+ B\omega^*||_2^2) + \frac{1}{\rho}\mathbf{E}(||\hat{\lambda}^{s-1} - \lambda^* - \lambda||_2^2 - ||\hat{\lambda}^{s} - \lambda^* - \lambda||_2^2)
  \end{aligned}
\end{equation}
Summing over $s=1,...,S$, we obtain:
\begin{equation}
\begin{aligned}
& 2KS\mathbf{E}(G(\bar{z}) + \langle\, \lambda, A\bar{x} + B\bar{\omega}\rangle)\\
& \leq \underbrace{\sum\limits_{s=1}^{S}\frac{1}{\eta_s}\mathbf{E} (||\hat{x}^{s-1} - x^*||_{\hat{G}^{s-1}}^2 - ||\hat{x}^s - x^*||_{\hat{G}^s}^2)}_{T_{15}} + \sum\limits_{s=1}^{S}  2\eta_sC_3K + \sum\limits_{s=1}^{S} K\frac{C_1(C_4+C_F)}{\sqrt{s}}\\
& + \rho||A\hat{x}^0 + B\omega^*||_2^2 + \frac{1}{\rho}||\hat{\lambda}^{0} - \lambda^* - \lambda||_2^2
\end{aligned}
\end{equation}
To bound $T_{15}$, recall $\hat{G}^{s-1} = \frac{1}{\sqrt{s}}I, \hat{G}^{s} = \frac{1}{\sqrt{s+1}}I$, then we have:
\begin{equation}
\begin{aligned}
  T_{15} &= \sum\limits_{s=1}^{S}\frac{1}{\eta_s}\mathbf{E} (||\hat{x}^{s-1} - x^*||_{\hat{G}^{s-1}}^2 - ||\hat{x}^s - x^*||_{\hat{G}^s}^2) \\
  &=\sum\limits_{s=1}^{S}\frac{1}{\sqrt{s}}\mathbf{E}||\hat{x}^{s-1} - x^*||_2^2 + \sum\limits_{s=1}^{S}\mathbf{E}(\sqrt{s}||\hat{x}^{s-1} - x^*||_2^2 - \sqrt{s+1}||\hat{x}^{s} - x^*||_2^2)\\
  & \leq 2\sqrt{S}D^2 + D^2
\end{aligned}
\end{equation}
Letting $\lambda = \Lambda\frac{A\bar{x} + B\bar{\omega}}{||A\bar{x} + B\bar{\omega}||}, \ \Lambda >0$ and using the same technique as in (\ref{eq:i4}), finally we obtain:
\begin{equation}
\begin{aligned}
  &\mathbf{E}(G(\bar{z}) + \Lambda||A\bar{x} + B\bar{\omega}||)\\
  & \leq \frac{C_1(C_4+C_F)}{\sqrt{S}} + \frac{D^2}{K\sqrt{S}}  + \frac{ C_3\log(S+1) }{S} + \frac{D^2+\rho||A^TA||D^2 + \frac{2}{\rho}||\hat{\lambda}^0 - \lambda^*||_2^2 +\frac{2}{\rho}\Lambda^2}{2KS}
\end{aligned}
\end{equation}
\QEDB



\end{document}